\documentclass{article}

\usepackage{graphicx}
\newcommand{\name}{\mbox{MITO}} 
\newcommand{\eqsize}{\footnotesize}
\usepackage{dsfont}
\usepackage{makecell}
\usepackage[normalem]{ulem}
\usepackage[dvipsnames]{xcolor}
\usepackage{pifont}
\usepackage[numbers]{natbib}
\usepackage[table]{xcolor}
\usepackage{boldline}
\usepackage[subtle]{savetrees}
\usepackage{wrapfig}

\definecolor{cvprblue}{rgb}{0.21,0.49,0.74}
\usepackage[pagebackref,breaklinks,colorlinks,allcolors=cvprblue]{hyperref}

\usepackage{enumitem}
\setlist[enumerate]{itemindent=12pt,
                    leftmargin=1pt, itemsep=1pt }
\setlist[itemize]{itemindent=12pt,
                    leftmargin=1pt, itemsep=1pt}
\usepackage{titlesec}
\titleformat{\section}
{\normalfont\Large\bfseries}{\thesection}{1em}{}
\titleformat{\subsection}
{\normalfont\large\bfseries}{\thesubsection}{1em}{}
\titleformat{\subsubsection}
{\normalfont\normalsize\bfseries}{\thesubsubsection}{1em}{}

\titlespacing{\section} {0pt}{0pt}{0pt}
\titlespacing{\subsection} {0pt}{0pt}{0pt}
\titlespacing{\subsubsection}{0pt}{0pt}{0pt}

 \usepackage[eandd, final]{neurips_2026}
\usepackage[utf8]{inputenc} % allow utf-8 input
\usepackage[T1]{fontenc}    % use 8-bit T1 fonts
\usepackage{hyperref}       % hyperlinks
\usepackage{url}            % simple URL typesetting
\usepackage{booktabs}       % professional-quality tables
\usepackage{amsfonts}       % blackboard math symbols
\usepackage{nicefrac}       % compact symbols for 1/2, etc.
\usepackage{microtype}      % microtypography
\usepackage{xcolor}         % colors
\usepackage{caption}
\usepackage{amsmath}
\title{\name: A Millimeter-Wave Dataset and Simulator for \\ Non-Line-of-Sight Perception}

\author{Tara Boroushaki$^{*1}$, Laura Dodds$^{*1}$, Cusuh Ham$^{2}$, Fadel Adib$^{1,3}$\\
tarab@mit.edu, ldodds@mit.edu, ham@adobe.com, fadel@mit.edu\\
$^1$ Massachusetts Institute of Technology, $^2$ Adobe Research, $^3$ Cartesian Syestems}

\begin{document}

\maketitle

\begin{abstract}
    The ability to observe the world is fundamental to reasoning and making informed decisions on how to interact with the environment. However, optical perception can often be disrupted due to common occurrences like occlusions, which can pose challenges to existing vision systems. We present MITO, the first millimeter-wave (mmWave) dataset of diverse, everyday objects, collected using a UR5 robotic arm with two mmWave radars operating at different frequencies and an RGB-D camera. Unlike visible light, mmWave signals can penetrate common occlusions (e.g., cardboard boxes, fabric, plastic). MITO captures over 24 million mmWave frames and uses them to generate 550 high-resolution mmWave images in line-of-sight and non-light-of-sight (NLOS), as well as RGB-D images, segmentation masks, and raw mmWave signals, taken from 76 different objects. We develop an open-source simulation tool that can be used to generate synthetic mmWave images for any 3D triangle mesh. Finally, we establish benchmarks for NLOS segmentation and classification to demonstrate the utility of our dataset and simulator for enabling broader NLOS perception.

\end{abstract}

\vspace{-0.1in}
\section{Introduction}
\vspace{-0.1pt}

Recent decades have witnessed many impressive advancements in computer vision, with state-of-the-art systems capable of high-accuracy object detection, segmentation, classification, and more. One of the main applications of these advances is in robotic manipulation and perception, where autonomous robots are expected to find and retrieve objects and navigate through complex environments spanning warehouses, manufacturing plants, and smart homes.

However, relying primarily on optical perception (e.g., cameras, LiDARs, etc) inherently limits existing systems to line-of-sight (LOS).\footnote{While some visible-light systems can rely on reflected laser light to image objects around the corner, they cannot image fully occluded items (e.g., within a closed box).} For example, a camera-based system cannot see inside a closed box to confirm e-commerce orders are correct or detect broken objects during shipping. Nor is it feasible for camera-based robots to efficiently plan complex, multi-step tasks in practical environments where required objects may be fully occluded. 

\begin{figure}
\centering
    \includegraphics[width=0.45\textwidth]{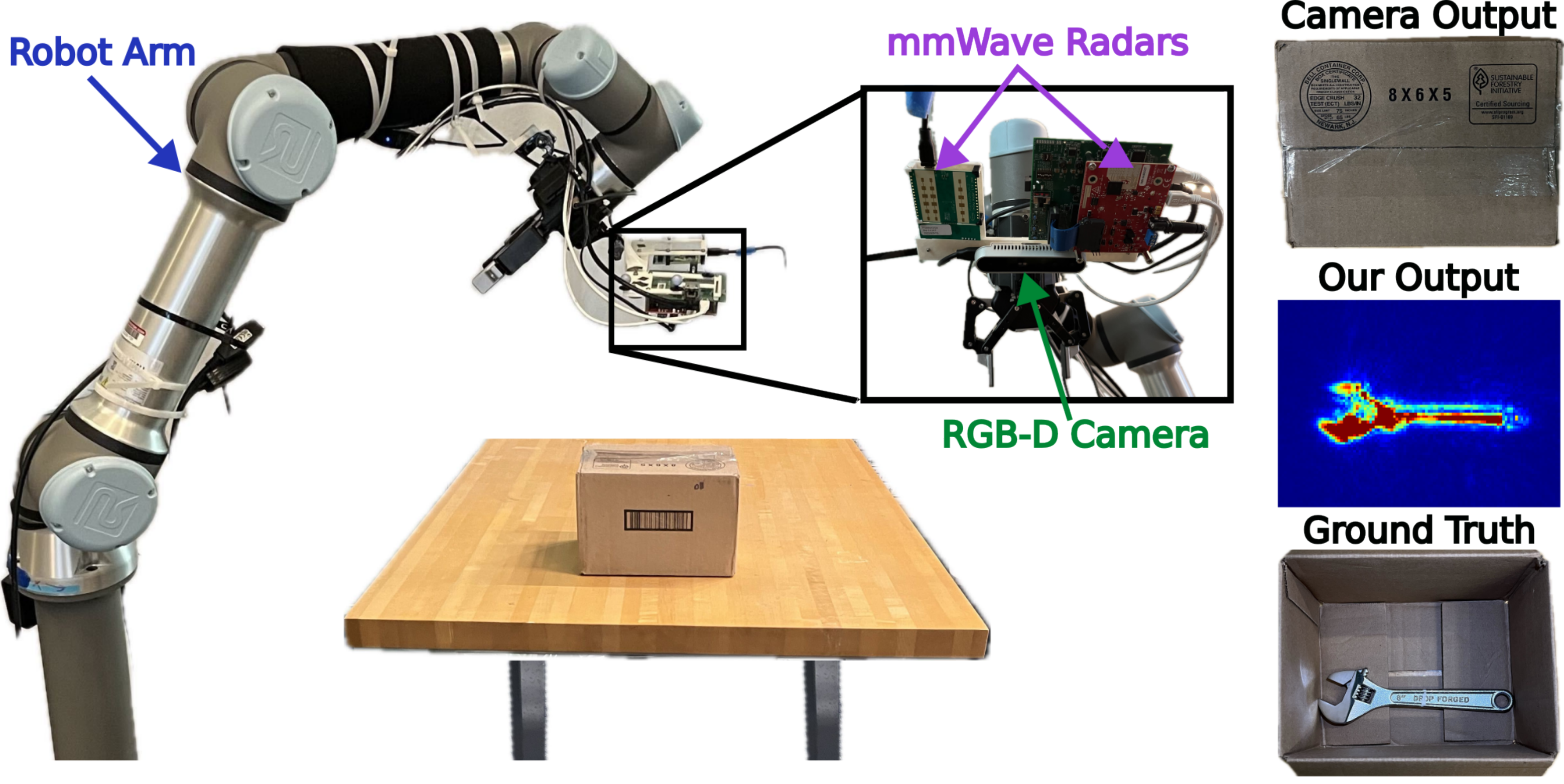}
    \caption{\footnotesize{\textbf{\name.}} \textnormal{We use a robotic arm to move mmWave radars throughout the environment. While an RGB-D camera's output cannot see inside the box, we produce high-resolution non-line-of-sight mmWave images. } }
    \label{fig:intro}
    \vspace{-0.2in}
\end{figure}

Instead, it is possible to produce non-line-of-sight (NLOS) images using millimeter-wave (mmWave) radars~\cite{ti_iwr1443}. Unlike visible light, mmWave wireless signals can traverse everyday occlusions such as cardboard, fabric, and plastic. This allows mmWave radars to image objects in non-line-of-sight, similar to how airport security scanners produce high-resolution NLOS images of passengers to detect hidden weapons. With the emergence of low-cost mmWave radars, it is possible to build pervasive autonomous systems with this NLOS imaging capability. For example, using mmWave radars attached to a robotic arm (Fig.~\ref{fig:intro}), we can produce a NLOS mmWave image of a wrench within a closed box.

However, to achieve the many possible applications of NLOS imaging, we need to be able to \textit{perceive} (e.g., segment, classify, etc) objects within mmWave images. Unfortunately, these images are fundamentally different from visible-light images in multiple ways: they 1) suffer from radar-specific artifacts (e.g., the much longer wavelength of mmWaves versus visible light--hundreds of nanometers--results in fundamentally different specular and diffuse properties of imaged objects), 2) have no color information, and 3) are complex-valued.\footnote{Interestingly, the values are dependent on not only the location of the object, but also the material and texture.} Therefore, since the majority of existing perception algorithms have not been trained on these mmWave images, 
they need to be adapted for NLOS perception. In some cases, we may require entirely new algorithms or models to enable mmWave-based perception to approach the accuracy of visual-light-based perception, as shown in this work. 

Inspired by how early datasets (e.g., MSRC~\cite{msrc_dataset}, Pascal~\cite{pascal_dataset}, Berkeley 3D~\cite{berkely_3d_dataset}, and YCB~\cite{ycb}) paved the way for many advancements within computer vision, we believe that a mmWave image dataset of everyday objects is fundamental for enabling pervasive NLOS perception. To summarize, we make the following key contributions: 

\begin{itemize}

    \item \textbf{Dataset.} We create MITO, a novel dataset of mmWave images, using a custom-built robotic imaging system and signal processing pipeline. We capture over 24 million mmWave frames, and use them to generate 550 images of 76 everyday objects in both line-of-sight and non-line-of-sight. For each object, we capture mmWave images at multiple different frequencies. We also provide RGB-D images of the objects (when in LOS) and ground truth object segmentations. 
    \item \textbf{Simulator.} We develop an open-source simulator that can produce synthetic mmWave images for any 3D triangle mesh. Since different materials interact differently with mmWave signals and, therefore, produce different images, we build two different modeling methods into our simulation, the combination of which can be used to simulate a wide range of material properties\footnote{The code and data will be added upon publication. A video demonstration is available here: 
\href{https://www.youtube.com/watch?v=atAqefGl3Sc}{https://www.youtube.com/watch?v=atAqefGl3Sc}}. 
    \item \textbf{Non-line-of-sight experiments. }Using \name, we establish baselines for two computer vision tasks in non-line-of-sight: 1)object segmentation, using the segment-anything model (SAM)~\cite{sam} and a mmWave power-based prompter, and 2) shape classification, which is trained entirely on synthetic images and generalizes to real images.
\end{itemize}

The dataset, code, and a video demonstration are available here: 

\href{https://github.com/signalkinetics/MITO\_Codebase}{https://github.com/signalkinetics/MITO\_Codebase} 

% \cusuh{Can't have anything identifying in the submission $\rightarrow$ remove + "will be made available" OR fully anonymize everything}

\vspace{-0.1pt}
\section{Related Work}
\vspace{-0.1pt}

\noindent \textbf{mmWave Datasets.}
Recognizing the importance of millimeter-wave perception, past research has made initial steps towards millimeter-wave datasets. However, the vast majority of them are focused on either imaging humans for action recognition~\cite{human_pose, milipoint, humans_and_cars,rahman2024mmvr, wu2024mmhpe}, outdoor imaging for autonomous driving~\cite{multimodal_av,av1,hawkeye_haitham,coloradar,kradar,radset,view-of-delft,yang2024autonomous, wang2024vision,guan2024talk2radar}, or imaging building interiors (walls, hallways)~\cite{brescia2024millinoise,prabhakara2023radarhd}. For higher-resolution imaging, datasets are limited to domain-specific objects like guns and knives for TSA where detecting concealed weapons through body scanners is important for airport security applications~\cite{human_concealed, concealed_class_1,SquiggleMilli}. Our work, however, is focused on generating a dataset of a larger, more diverse set of everyday objects for applications such as robotic manipulation and perception.

\vspace{0.03in}
\noindent \textbf{mmWave Image Simulators.}
Given the growth in millimeter-wave devices for networking and sensing, the vast majority of existing mmWave simulations have also focused on understanding channel characteristics (e.g., for 5G networks)~\cite{ansys,ChanSim2,ChanSim1, sionna}  or for imaging humans for action recognition and HCI applications~\cite{hsim_1,hsim_2,hsim_3}.
\textcolor{black}{However, \cite{ChanSim2,ChanSim1, sionna} do not support small indoor objects, and \cite{hsim_1,hsim_2,hsim_3} focus exclusively on human body modeling, rely on assumptions about scattering from limbs (which do not generalize to arbitrary objects). In principle, simulators such as \cite{ansys, SquiggleMilli}} could be used for generating mmWave images of everyday objects, but they are extremely expensive, costing tens of thousands of dollars~\cite{ansys} and/or are specific to simulating metallic items, limiting their application to everyday objects with different properties~\cite{SquiggleMilli}.

In contrast, we develop an open-source simulator and demonstrated that it generates synthetic data that matches real-world collected data. We believe that this tool, combined with advances in vision algorithms, can lead to improvements in performance in non-line-of-sight scenarios.

\vspace{0.03in}
\noindent \textbf{mmWave Perception.}
Prior work has demonstrated the use of millimeter waves for various perception tasks, including detection and classification. However, these approaches are limited to specific domains, such as vehicles~\cite{human_detect_av, hawkeye_haitham, car_completion}, pedestrians~\cite{ped_classification}, weapons~\cite{concealed_class_1, concealed_class_2}, or entirely metallic objects~\cite{obj_in_box}. Through a new dataset and simulator, our work aims to expand the application of mmWave perception to a more open domain consisting of diverse categories.

\vspace{-0.1pt}
\section{Background}
\vspace{-0.1pt}

\begin{figure}[t]
\begin{minipage}[t]{0.23\textwidth}
\centering
    \includegraphics[width=\textwidth]{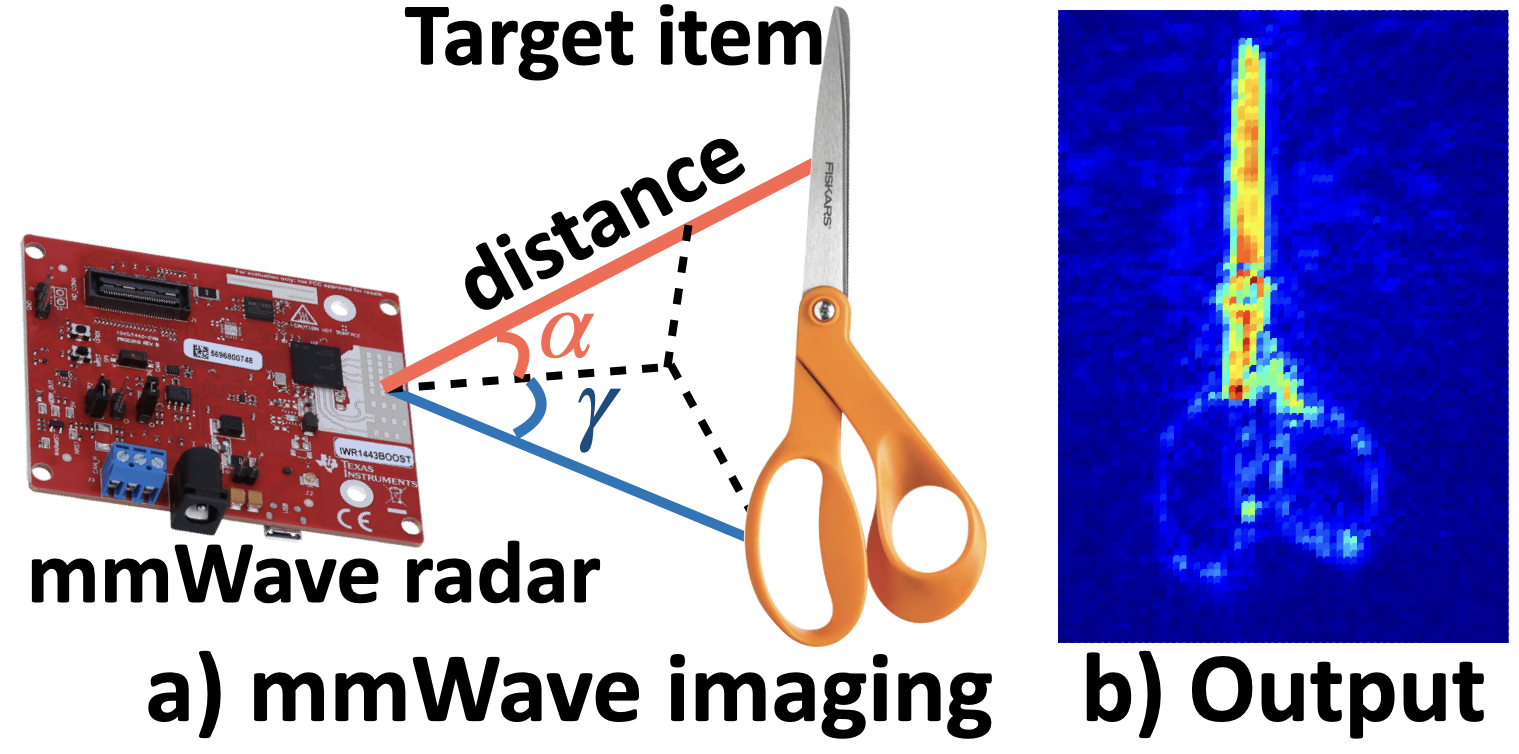}
    \vspace{-0.23in}
    \caption{\footnotesize{\textbf{mmWave Imaging}} \textnormal{a) mmWave radars  estimate range and angle-of-arrival to produce b) reflection maps.}} 
    \label{fig:aoa}
    \vspace{-0.225in}
\end{minipage}
\hfill
\begin{minipage}[t]{0.23\textwidth}
\centering
        \includegraphics[width=\textwidth]{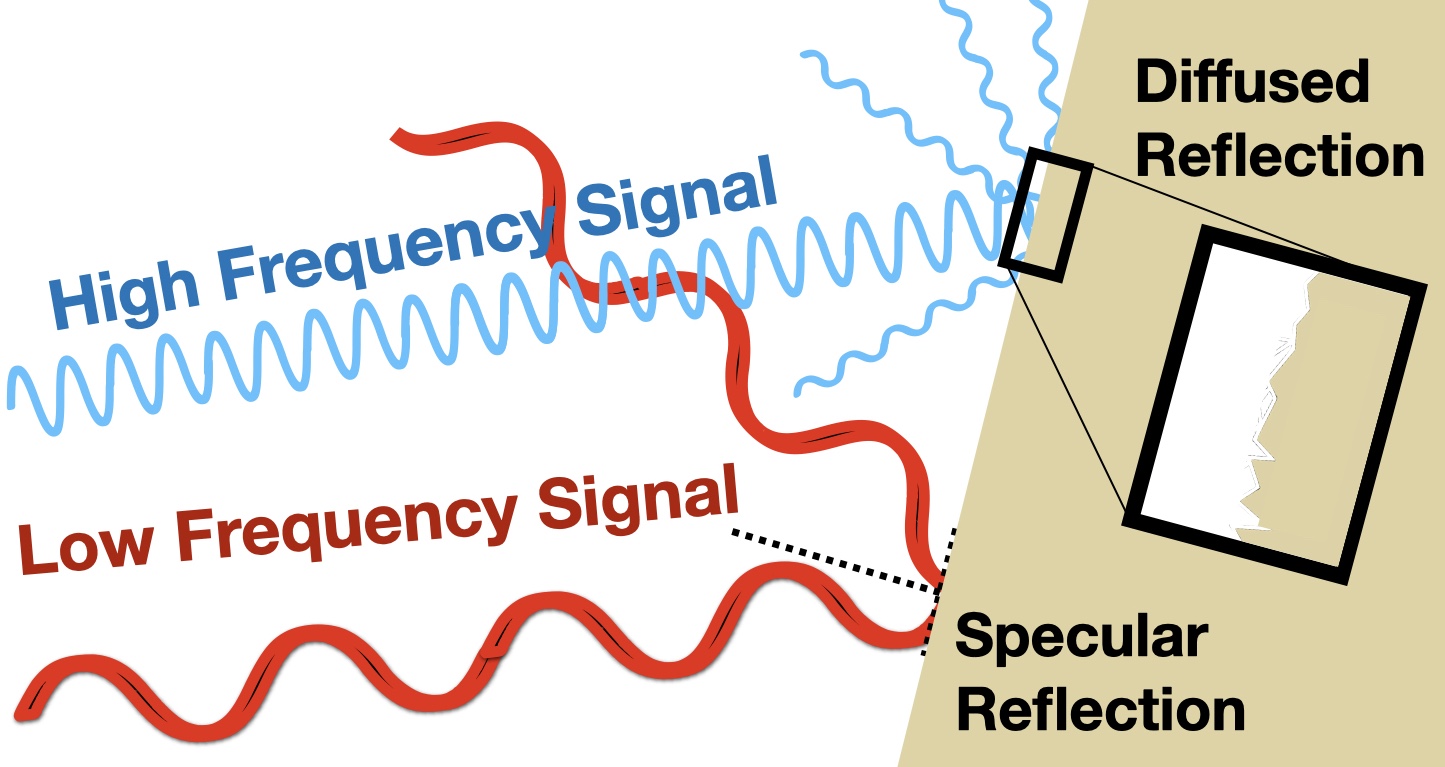}
    \vspace{-0.23in}
    \caption{\footnotesize{\textbf{Diffuse and Specular Reflections.}} \textnormal{The frequency of a signal changes the type of reflections it experiences for the same surface.}} 
    \label{fig:diffuse}
    \vspace{-0.225in}
\end{minipage}
\hfill
\begin{minipage}[t]{0.5\textwidth}
\fontsize{8pt}{6pt}\selectfont
\setlength{\tabcolsep}{0.2pt}
\setlength\extrarowheight{2pt}
\centering
\begin{tabular}{ |c|c|c|c|c| }
 \hline \textbf{Dataset} &\textbf{Targets} &\textbf{FCC}  & \textbf{Publicly} & \textbf{Resolution}  \\
 & &\textbf{Band-} &\textbf{Available}  &  \\  
 & &\textbf{width}  &&  \\  
 \hline
 % \Xhline{2\arrayrulewidth}
 \textbf{MITO}   & \textcolor{ForestGreen}{76 Everyday} &\textcolor{ForestGreen}{\checkmark}  &\textcolor{ForestGreen}{\checkmark} & \textcolor{ForestGreen}{High}  \\  
 & \textcolor{ForestGreen}{Objects} &  & &  \\  
 \hline
 % \Xhline{2\arrayrulewidth}
\cite{human_concealed} & \textcolor{brown}{30 TSA Objects} & \textcolor{red}{\ding{55}} & \textcolor{red}{\ding{55}} & \textcolor{ForestGreen}{High}  \\  
  \hline
 \cite{SquiggleMilli}  &  \textcolor{red}{9 Objects} & \textcolor{red}{\ding{55}} &\textcolor{ForestGreen}{\checkmark} & \textcolor{ForestGreen}{High}    \\  
  &  \textcolor{red}{Categories} &  &  &     \\  
  \hline
 \cite{human_pose, humans_and_cars, milipoint, rahman2024mmvr, wu2024mmhpe}  &  \textcolor{red}{Humans} &\textcolor{ForestGreen}{\checkmark}  &\textcolor{ForestGreen}{\checkmark}& \textcolor{red}{Low}  \\  
  \hline
 \cite{multimodal_av,av1,hawkeye_haitham,coloradar}   &  \textcolor{red}{Vehicles \&} &  &  &   \\ 
  % \hline
 \cite{radset,view-of-delft,yang2024autonomous,guan2024talk2radar}  &  \textcolor{red}{Driving Scenes} &\textcolor{ForestGreen}{\checkmark} &\textcolor{ForestGreen}{\checkmark} & \textcolor{red}{Low}  \\ 
   \hline
% \hline
\end{tabular}
\vspace{-0.1in}
    \captionof{table}{\footnotesize{\textbf{mmWave Dataset Comparisons. }} }
    \label{tab:datasets}
    \vspace{-0.15in}
\end{minipage}
\end{figure}

\noindent \textbf{mmWave Imaging.} \label{sec:bg_imaging}
A millimeter-wave radar transmits wireless signals with mmWave wavelengths, captures their reflections, and uses them to generate 3D images (shown in Fig.~\ref{fig:aoa}a). It uses two core signal processing techniques. First, the radar estimates the distance of each reflection by comparing the frequency between the transmitted and received signals. Second, it measures the angle of reflection (in both azimuth and elevation) by comparing the received signals between different antennas. 3D images can then be formed by calculating the reflection power emanating from each voxel in 3D space. 

As shown in Fig.~\ref{fig:aoa}b, the captured image can be represented as a heatmap, where red represents high intensity reflection and blue represents low intensity. It is important to note that this visualization is a 2D projection of the 3D image. Mathematically, since every sample from the radar is a complex number with a phase and an amplitude, we can combine them with the following equation: 
{\vspace{-0.05in}\eqsize
\begin{equation}
    I(x,y,z) = \sum_{k=1}^{K} \sum_{j=1}^{N} S_{j,k} e^{\frac{j 2 \pi d(x,y,z,k)}{\lambda_j}}
\label{eq:sar}
\end{equation}
}
\noindent where $I(x,y,z)$ is the complex mmWave image at location $(x,y,z)$, $K$ is the number of antennas, $N$ is the number of samples in each signal, $S_{j,k}$ is the j\textsuperscript{th} sample of the received signal from the k\textsuperscript{th} antenna, and  $\lambda_j$ is the wavelength of the signal at sample j, which is proportional to the speed of light c and the frequency f as $\lambda = \frac{c}{f}$. $d(x,y,z,k)=2||(x,y,z)-p_k||$ is the round-trip distance from the k\textsuperscript{th} antenna (at location $p_k$) to the point $(x,y,z)$.

\noindent \textit{Resolution.} One important characteristic of a mmWave image is its resolution. The resolution in range (depth) is determined by the \textit{bandwidth}, or the range of frequencies that the signal covers. We compute the depth resolution     $\delta_z = \frac{c}{2B}$
 where $c$ is the speed of light, and $B$ is the bandwidth of the signal. The resolution in the horizontal and vertical dimensions is dependent on the \textit{aperture}, or the distance between the first and last antenna in that dimension.
$ \delta_x = \frac{\lambda z_0}{2 D_x}$ and $\delta_y = \frac{\lambda z_0}{2 D_y}$ are the resolution in the x and y dimensions, respectively. $z_0$ is the range to the target, and $D_x$ and $D_y$ are the apertures in x and y dimensions, respectively.

\noindent\textit{Synthetic Aperture Radar (SAR) Imaging.} As described above, it is desirable to increase the aperture to increase the image resolution. Instead of creating a large aperture through a large number of physical antennas on the radar, it is possible to perform SAR imaging. In this method, one set of transmitters/receivers are moved through the environment, taking measurements from different locations, and constructing a ``synthetic aperture". These measurements can be combined through Eq.~\ref{eq:sar} to provide the same resolution as a physical aperture.

% \vspace{-0.08pt}
\noindent \textbf{Impact of Frequency.} \label{sec:bg_freq} 
The frequency used for imaging introduces an important tradeoff -- higher frequencies result in higher resolution, but the signals experience more \textit{attenuation}, or power loss, as they travel through materials~\cite{textbook_attenuation}. 
In other words, high frequency signals (e.g., Terahertz) struggle with thick occlusions due to higher noise~\cite{thz_penetration}, but lower frequencies (e.g., WiFi) suffer from poor resolution~\cite{adib20143d}. Therefore, we choose multi-spectral (multi-frequency) mmWave signals for our dataset and simulator. Specifically, we select 24~GHz and 77~GHz since these are the bands approved by the FCC for industrial and commercial use~\cite{fcc_reg}, and most commercial mmWave radars operate within them. 

% \vspace{-0.01in}
\noindent \textbf{Types of Reflections.} \label{sec:bg_refl}
% \vspace{-0.03in}
mmWave signals primarily exhibit specular (mirror-like) reflections~\cite{mmwave_specular}, illustrated in Fig.~\ref{fig:diffuse}. Due to the significantly longer wavelengths of mmWaves compared to visible light (millimeters vs. nanometers), surfaces that appear diffuse (i.e., omnidirectional) for visible light (blue) often appear specular for mmWave signals (red). mmWaves also exhibit edge diffraction, or strong reflections when a wave strikes a sharp edge, scattering it in multiple directions. Edge diffraction can impact signal propagation and perception differently from diffuse or specular reflections. These reflection properties are one fundamental difference between mmWave signals and visible light that necessitates models and algorithms specifically designed for mmWave images.

\vspace{-0.1in}
\section{\name\ Dataset}
\label{sec:dataset}
\vspace{-0.1in}

We describe the process of creating \name, which contains over 24 million mmWave frames where each frame is a complex-valued time series. These frames are used to generate 550 unique, high-resolution 3D mmWave SAR images of 76 diverse everyday objects. \name~also includes multi-spectral (77~GHz and 24~GHz) mmWave images, RGB-D images, and ground truth object segmentation masks.
Table~\ref{tab:datasets} provides an overview comparing \name\ with other existing mmWave datasets.

\noindent \textbf{Object Selection.}
% \vspace{-0.03in}
\name\ builds on the YCB dataset~\cite{ycb}, a standard dataset of everyday objects used for robotic manipulation. We select 76 out of 80 objects from the YCB dataset (the remaining 4 were discarded due to not being able to obtain the exact object). Unlike previous mmWave datasets, these objects cover a diverse range of shapes, sizes, materials, and categories (additional details on object diversity  can be found in the supplementary). The YCB dataset also provides 2D images and 3D object meshes, which we use to simulate mmWave images. 

\begin{figure*}
\begin{minipage}[t]{0.7\linewidth}
    \includegraphics[width=\textwidth]{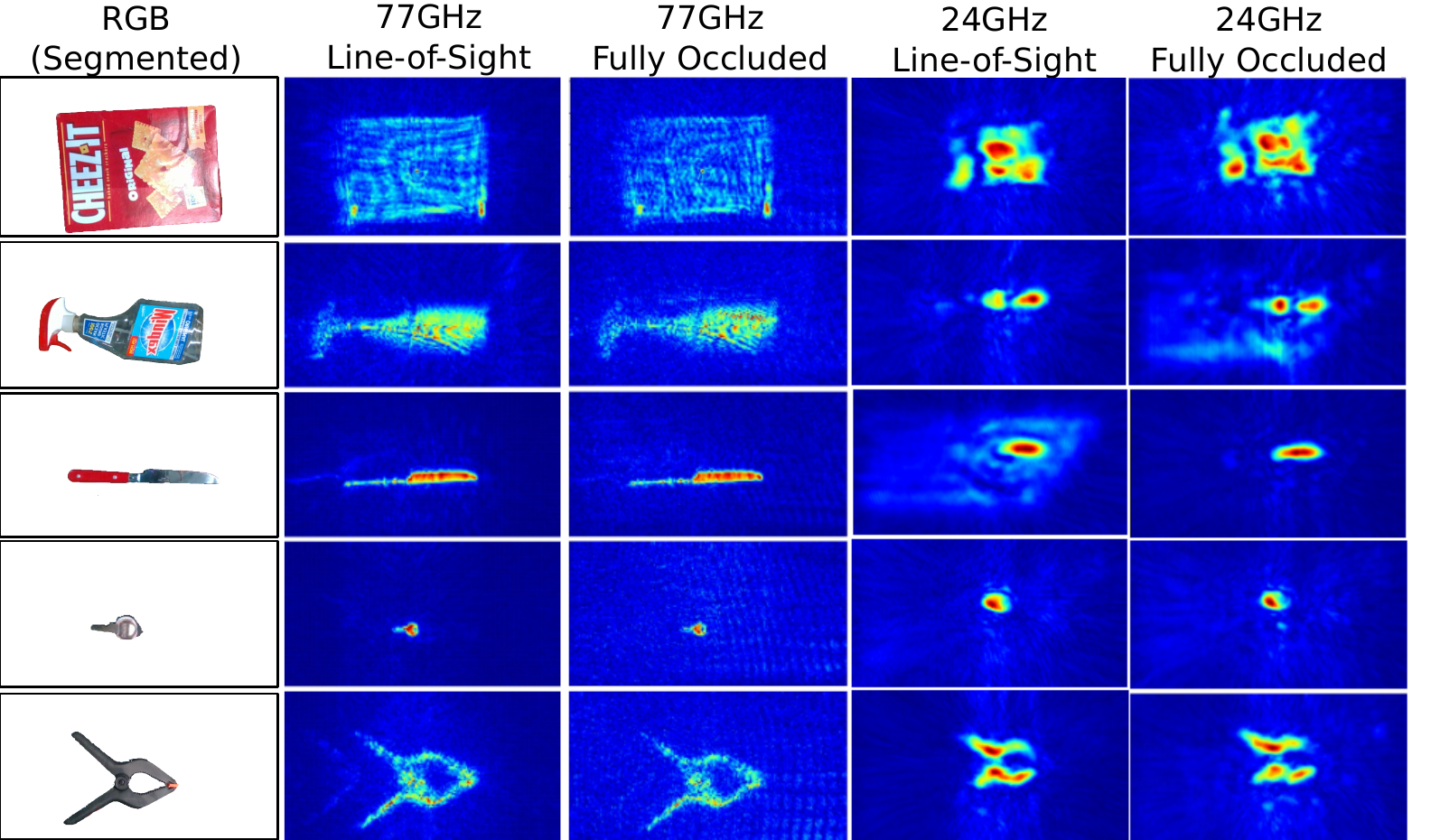}
    \vspace{-0.2in}
    \caption{\footnotesize{\textbf{Sample Images.}} \textnormal{Example mmWave images from \name.}  }
    \label{fig:all_img}
    \end{minipage}
    \hfill
\begin{minipage}[t]{0.29\linewidth}
    
    \vspace{-2.1in}
    \includegraphics[width=\linewidth]{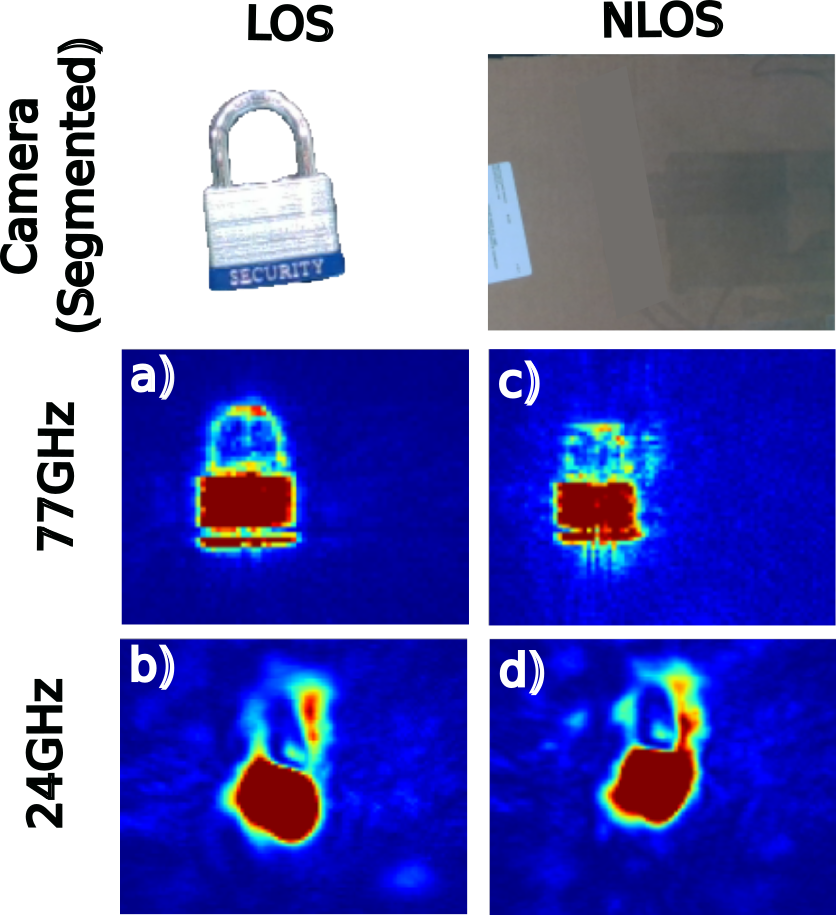}
    \vspace{-0.2in}
    \caption{\footnotesize{\textbf{Multi-Spectral Imaging.}} \textnormal{a-b) show 77~GHz \& 24~GHz images for a padlock in LOS. c-d) show the images in NLOS. } }
    \label{fig:multispectral}
    % \vspace{-0.25in}

\end{minipage}
\vspace{-0.25in}
\end{figure*}

% \vspace{-0.03in}
\noindent \textbf{Data Collection Protocol.}
% \vspace{-0.03in}
\noindent\textit{Real-World mmWave Images.} Fig.~\ref{fig:intro} visualizes our data collection setup. We connect a 77~GHz radar (TI IWR1443Boost~\cite{ti_iwr1443} and DCA1000EVM~\cite{dca1000evm}), 24~GHz radar (Infineon Position2Go~\cite{position2go}), and Intel Realsense D415~\cite{realsense} to an end-effector of a UR5e robotic arm~\cite{UR5} using custom-designed 3D-printed parts. Since the robot and radars are controlled from different computers, we synchronize them by setting both computers to the same NTP time server.

We place the objects in two setups that minimize background reflections in different ways. In the first, we fix each object onto a tripod by hot-gluing a 3D-printed nut to each object and screwing it onto a threaded tripod. Because there is no surface directly below the object, any strong reflections from the background are avoided. In the second, we place the object directly on a styrofoam block, since styrofoam is largely transparent to mmWave signals~\cite{ti_styrofoam}. We also include experiments where the tripod/styrofoam is placed in the scene without any objects so that the impact of the tripod/styrofoam can be removed by subtracting the empty image from the main image~\cite{RFCapture}. 

At the start of an experiment, the robot moves the camera to the middle of the workspace to take an RGB-D image prior to any potential occlusion (i.e., in the NLOS scenario). Then, it moves the radars throughout the workspace, covering a rectangular space. The aperture of our experiments range from 55cm$\times$22cm to 60cm$\times$45cm, ensuring that each experiment uses an aperture which covers the full object. The radars continuously measure mmWave signals while the robot is moving, and the robot records its trajectory over time. After we complete an experiment in LOS, we run an experiment in NLOS without moving the object. We place a layer of cardboard over the object to simulate the case where the object was placed inside a box. We also collect experiments with four layers of fabric over the cardboard to simulate more dense occlusions.

% \begin{wrapfigure}{r}{0.45\textwidth}
% \begin{minipage}[t]{0.4\linewidth}
\begin{table}[t]
% \vspace{-0.251in}
% \scriptsize
\fontsize{7.5pt}{5.5pt}\selectfont
\setlength{\tabcolsep}{5pt}
\setlength\extrarowheight{3pt}
\centering
\begin{tabular}{ |c|c|c|c|c|c| }
 \hline \textbf{Simulator} &\textbf{Compatible Objects} &\textbf{Simulates Objects \& Indoors}  & \textbf{Material Diversity}& \textbf{Publicly Available} &  \textbf{Cost}\\
 % & \textbf{Objects} &\textbf{Objects \&} & \textbf{Diversity} & \textbf{Available} & \\  
 % & &\textbf{Indoors}  && \textbf{ } & \\  
 \hline
 % \Xhline{2\arrayrulewidth}
 \textbf{MITO}   & \textcolor{ForestGreen}{Everyday Objects} & \textcolor{ForestGreen}{\checkmark}  & \textcolor{ForestGreen}{\checkmark} & \textcolor{ForestGreen}{\checkmark} & \textcolor{ForestGreen}{Free} \\  
 \hline
 % \Xhline{2\arrayrulewidth}
 Sionna\cite{sionna}   &  \textcolor{brown}{Cars, streets, buildings, etc} & \textcolor{red}{\ding{55}} & \textcolor{ForestGreen}{\checkmark} & \textcolor{ForestGreen}{\checkmark} & \textcolor{ForestGreen}{Free} \\   
  \hline
Ansys~\cite{ansys}  &  \textcolor{ForestGreen}{Everything} & \textcolor{ForestGreen}{\checkmark} & \textcolor{ForestGreen}{\checkmark} & \textcolor{red}{\ding{55}}  & \textcolor{red}{10-50K\$}   \\  
  \hline
\cite{SquiggleMilli}  &  \textcolor{red}{ Metalic objects} & \textcolor{ForestGreen}{\checkmark} & \textcolor{red}{\ding{55}} & \textcolor{red}{\ding{55}}  & \textcolor{ForestGreen}{Free}  \\  
  \hline
 \cite{hsim_1} &  \textcolor{red}{Humans} & \textcolor{red}{\ding{55}}  & \textcolor{red}{\ding{55}}& \textcolor{red}{\ding{55}}&  {--}  \\  
  \hline
 \cite{hsim_2}  &  \textcolor{red}{Humans} & \textcolor{red}{\ding{55}} & \textcolor{red}{\ding{55}} & \textcolor{ForestGreen}{\checkmark}& \textcolor{ForestGreen}{Free}  \\ 
  \hline
 \cite{hsim_3}  &  \textcolor{red}{Humans} & \textcolor{red}{\ding{55}} & \textcolor{red}{\ding{55}} & \textcolor{ForestGreen}{\checkmark}& \textcolor{ForestGreen}{Free}  \\ 
   \hline
\end{tabular}
% \vspace{-0.05in}
    \captionof{table}{\footnotesize{\textbf{mmWave Simulators Comparisons. }}}
    \label{tab:sim_comparison}
    \vspace{-0.3in}
\end{table}
% \end{minipage}
% \end{wrapfigure}

\noindent\textit{Synthetic mmWave Images.} In addition to the real-world data, we also provide a simulation tool to generate synthetic mmWave images, which can be used to increase the volume of data to aid in model training. We include at least 2 simulation images per object, along with the simulator for producing additional images. % (Sec.~\ref{sec:simulation}). 

% \vspace{-0.25pt}
\noindent \textbf{Data Processing.}
% \vspace{-0.24pt}
To process a real-world mmWave image, we need to know the location where each measurement was taken. Thus, we interpolate the timestamped robot locations to match the timestamp of the radar measurements. The interpolated locations, in addition to the raw radar measurements, are then processed into a 3D mmWave image by applying Eq.~\ref{eq:sar}. Example images can be seen in Fig.~\ref{fig:all_img}, which shows the segmented RGB images (1st column), 77~GHz LOS (2nd column), 77~GHz NLOS (3rd column), 24~GHz LOS (4th column), and 24~GHz NLOS (5th column) mmWave images. {Additional examples, including an analysis of the impact of various occlusion types (such as plastic, drywall, and plywood), are provided in the supplementary materials.}%Additional images are provided in the supplementary.

We produce 550 3D, complex-valued mmWave images at two different frequencies (24~GHz \& 77~GHz) in both LOS and NLOS settings using 24 million raw measurements and locations.  We include the raw data in \name~ to enable flexibility in leveraging and/or postprocessing the data using a custom pipeline. We also include images of objects at different angles and images with multiple objects.

For each mmWave image, we use its corresponding RGB-D image (in LOS) to provide a ground truth object segmentation mask. The segmentation masks are produced by using an interactive interface for the segment-anything model (SAM)~\cite{sam} to select the points on the RGB picture as a prompt for the model. The human annotator continues selecting points until the mask covers the full object. We then use the camera's location to transform and align the RGB-D ground-truth mask with the mmWave image. In addition, we align the 3D ground-truth mesh model of each object with the corresponding depth images. This alignment enables richer annotations beyond segmentation, including 3D bounding boxes and full 6-DoF object pose labels. More details on the alignment process are in the supplementary.

% \vspace{-0.1pt}
\noindent \textbf{Analysis of Different Frequencies.}
% \vspace{-0.1pt}
% \textcolor{green}{Ta: should we move this to suppmat?}
Recall from Sec.~\ref{sec:bg_freq} that the frequency of the mmWave signal impacts both the resolution of the resulting image and the ability for the signal to traverse through occlusions. 
We analyze this phenomenon in real-world images in \name\ in Fig.~\ref{fig:multispectral}. Fig.\ref{fig:multispectral}a-b show a 77~GHz and 24~GHz image of a padlock in LOS, respectively. The 77~GHz image is higher resolution, as expected. On the other hand, Fig.~\ref{fig:multispectral}c-d show 77~GHz and 24~GHz images of the same object in NLOS (placed underneath cardboard). In this case, the 24~GHz image is very similar to the LOS case, while the 77~GHz shows some artifacts that were not present in LOS. This is because the 24~GHz signal travels through the cardboard relatively unaffected, while the 77~GHz is partially reflected and absorbed. We hope that by including two different frequencies, future algorithms can leverage both images to benefit from high-resolution and robustness to occlusions.

\vspace{-0.1in}
\section{\name\ Simulator}\label{sec:simulation}
\vspace{-0.05in}

We describe our \name~mmWave image simulator. Unlike other mmWave simulators, \name\ is compatible with diverse objects, accommodates variable material properties, and is open-source. We provide a comparison to other mmWave simulators in Table~\ref{tab:sim_comparison}.

\vspace{-0.1in}
\subsection{mmWave Simulation}
\vspace{-0.05in}

Our simulator takes as input a 3D triangle mesh and radar locations to simulate mmWave signals from. The mesh files can be sourced from object datasets (e.g., YCB~\cite{ycb}), online repositories (e.g., 3D Warehouse, Turbo Squid), or custom-generated on smartphones (e.g., PolyCam iPhone app).

At a high level, our simulator works by estimating the raw signals expected at each of these locations, given reflections off the target object. Then, synthetic images are produced by feeding these simulated signals into the same image processing pipeline used for real world data (Eq.~\ref{eq:sar}).

The process of estimating raw signals at a given location is as follows. (1) We find which mesh vertices in the mesh are visible from that location~\cite{mesh_visibility}. (2) For a given visible vertex, we measure the distance to the measurement location. Given the distance, we simulate the reflection from this vertex as:{ \eqsize
    \begin{equation} \label{eq:single_sim}
        s_j(l,v) = e^{-j \frac{4 \pi |l-v|}{\lambda_j}}
    \end{equation}
    }
\noindent where $s_j(l,v)$ is the j\textsuperscript{th} sample of the reflection from  location $l$ to vertex $v$ and back. (3) We sum the simulated reflections from all visible vertices $V$ to produce the raw signal for this measurement location: $ S_j = \sum_{v\in V} s_j(l,v) $ where $S_j$ is the j\textsuperscript{th} sample of the simulated signal from  location $l$.

\noindent After repeating this for each measurement location, we feed the raw signals through our imaging processing pipeline to produce the final synthetic image. Note that we can rotate the mesh files to produce synthetic images at various angles. Additionally, recall from Sec.~\ref{sec:bg_refl} that the object's material and texture may result in different types of reflections. To accurately model different objects, we introduce two variations of the simulation to account for different types of reflections.

\noindent\textbf{Specular Reflections.} When an object exhibits purely specular reflections, the signal will reflect off each vertex in only one direction. In some cases, the signal is reflected away from the radar's receivers, and therefore should not be added to $S_j$. Specifically, the signal will only be reflected back towards the radar if the surface normal is pointing towards the radar's location. Otherwise, the signal will be reflected away.

We update our above simulation to account for this by removing reflections where the angle between the normal ($\boldsymbol{n}$) and the vector from the radar to the vertex ($l-v$) is less than a threshold $\tau$. Formally: %, we can update Eq.~\ref{eq:single_sim} to:
%
%
% \vspace{-0.1pt}
{ \eqsize
\begin{equation}
    s_j(l,v) =  \begin{cases} 
      0 & \cos^{-1}(\frac{\boldsymbol{n} \cdot (l-v)}{|\boldsymbol{n}| |(l-v)|}) \geq \tau \\
      e^{-j \frac{4 \pi |l-v|}{\lambda_j}} & \cos^{-1}(\frac{\boldsymbol{n} \cdot (l-v)}{||\boldsymbol{n}| |(l-v)|}) < \tau
   \end{cases}
\end{equation}}

\noindent Fig.~\ref{fig:sim}b shows an example when simulating entirely based on specular reflections.

\noindent\textbf{Edge Reflections.} In other cases, the signal experiences scattering effects from the edges of objects.
First, we define a vertex as being on an edge of an object when the angle between two of its faces is greater than an angle $\tau_e$. We find the set of all vertices that lie on an edge, $V_e$, and replace Eq.~\ref{eq:single_sim} such that it only allows reflections from these vertices:

\vspace{-0.1pt}
{ \eqsize
\begin{equation}
    s_j(l,v) =  \begin{cases} 
      0 & v \notin V_e \\
      e^{-j \frac{4 \pi |l-v|}{\lambda_j}} & v \in V_e
   \end{cases}
\end{equation}}

\noindent Fig.~\ref{fig:sim}c shows an example when simulating edge reflections.

\noindent\textbf{Combining Reflections.} In practice, materials rarely exhibit only one type of reflection, but rather some combination of the two. To account for this, our synthetic images can be added together with different weights to simulate different materials:

\begin{figure*}
\begin{minipage}[t]{0.73\linewidth}
\centering
    \includegraphics[width=1\textwidth]{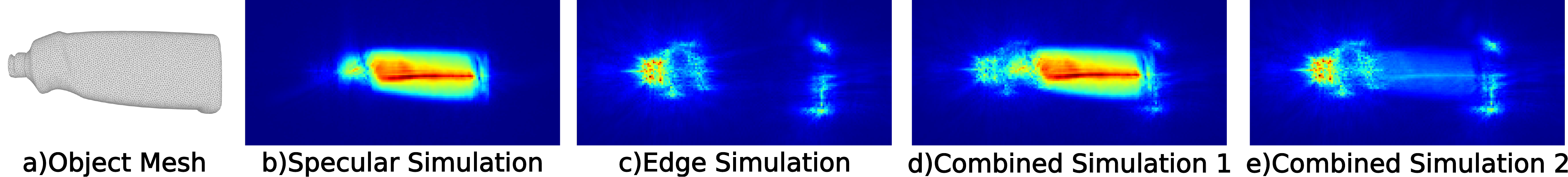}
    \vspace{-0.25in}
    \caption{\footnotesize{\textbf{Simulation Output.}} \textnormal{Our simulation uses a) a 3D mesh to produce images assuming different reflection types: b) specular \& c) edge. d/e) We combine these with different weights to simulate different materials.}}
    \label{fig:sim}
    \vspace{-0.2in}
\end{minipage}
\hfill
\begin{minipage}[t]{0.26\linewidth}
\label{table:sim}
\vspace{-0.6in}
\footnotesize
\label{table:sim}
\setlength{\tabcolsep}{4pt}
\begin{tabular}{ |c|c|c|c| }
 \hline \textbf{Sim} &\textbf{25\textsuperscript{th}} &\textbf{50\textsuperscript{th}} &\textbf{75\textsuperscript{th}}     \\  
    \Xhline{3.5\arrayrulewidth} %\hline \hline
 Comb   &  \textbf{85\%} & \textbf{94\%} & \textbf{98\%}  \\  \Xhline{3.5\arrayrulewidth} %\hline \hline
 Spec & 68\% & 90\% & 97\%\\ \hline
    Edge  & 49\% & 71\%  & 88\%\\\hline
\end{tabular}
\vspace{-0.08in}
\captionof{table}{\footnotesize{\textbf{Simulation 3D F-Score} \textnormal{for combined (Comb), specular (Spec), \& edge simulations.}}}
\vspace{-0.2in}
\label{table:sim}
\end{minipage}
\end{figure*}

% \vspace{-0.1pt}
{ \eqsize
\begin{equation}
    I_{syn}(\alpha_1, \alpha_2) = \frac{\alpha_1}{\alpha_1 + \alpha_2} I_s + \frac{\alpha_2}{\alpha_1 + \alpha_2} I_e
\end{equation}}

\noindent where $I_{syn}(\alpha_1, \alpha_2)$ is the combined synthetic image for weights $\alpha_1, \alpha_2$, and $I_s$ and $I_e$ are the images from the specular and edge simulations, respectively.
Fig.~\ref{fig:sim}d-e shows two examples of combining simulations with different weights representing two different materials.

\vspace{-0.5pt}
\subsection{Simulation Accuracy}
\vspace{-0.1pt}
To evaluate the accuracy of our simulator, we use the standard 3D F-score metric~\cite{3d_fscore} to compare the mmWave outputs from real-world and simulation, defined as:
\vspace{-0.05in}
{\eqsize
\begin{equation}
    PR = \frac{1}{N_R} \sum_{i=1}^{N_R} \mathds{1}_{d(P_{R}(i),P_{S})<\tau_F}, \hspace{0.1in}
    RE = \frac{1}{N_{S}} \sum_{j=1}^{N_{S}} \mathds{1}_{d(P_S(j),P_{R})<\tau_F}, \hspace{0.1in}
    F = \frac{2\ PR \ RE}{PR+RE} 
\end{equation}
}
\vspace{-0.05in}
% {\eqsize
% \begin{equation}
% \begin{aligned}
%     RE &= \frac{1}{N_{S}} \sum_{j=1}^{N_{S}} \mathds{1}_{d(P_S(j),P_{R})<\tau_F}, \hspace{0.1in}
%     F = \frac{2\ PR \ RE}{PR+RE}  
% \end{aligned}
% \end{equation}
% }
% \vspace{-0.05in}
%

\noindent where $PR$, $RE$, and $F$ are the precision, recall, and F-score. $\mathds{1}$ is an indicator variable. $N_R$ ($N_S$) is the number of points in the real (synthetic) point cloud. ${\small \tau_F}$ is the F-score distance threshold. $d(x, P)$ is the distance from point x to its nearest neighbor in cloud P: ${\footnotesize d(x, P) = \min\limits_{x'\in P} ||x-x'||}$.

    Since our goal is not to select a single set of weights for each object, but to allow the simulation to represent many different objects, we compute this metric across a range of weights $\{ \alpha_1, \alpha_2\}$ and choose the combined simulation which produces the best F-score (e.g., the simulation which best matches the properties of this real-world object). The F-score for a synthetic image $I_{syn}$ and real-world image $I$ is $F = \max_{(\alpha_1, \alpha_2) \in W} F(I_{syn}(\alpha_1, \alpha_2), I)$ where $W$ is the set of all weights.

\noindent\textbf{Aligning the Point Clouds.}
Before we compute the 3D F-score, we first need to convert both the real-world and simulation images to a point cloud. To do so, we apply a standard mmWave point cloud generation procedure\cite{LauraEPFL} by
creating one point in the center of each mmWave voxel that has a power above a threshold $\tau_P$.i.e.,  $P_R = \{ p \mid |I(p)| > \tau_P \}$ and  $P_S = \{ p \mid |I_{\text{syn}}(p)| > \tau_P \}$, 
where $P_R$ and $P_S$ are the set of points created for the real and synthetic mmWave images, respectively. Next, we align the real-world and simulation point clouds using a standard iterative closest point implementation~\cite{icp}. This is necessary since our simulated objects are in the center of the image, while our real world experiments may have objects offset from the center.

\noindent\textbf{Results.} We use the above process to compute the 3D F-score for all objects in the dataset.
We also compare the accuracy when using only specular simulation images (i.e., $\alpha_1=1, \alpha_2=0$), and when using only edge simulation images (i.e., $\alpha_1=0, \alpha_2=1$). 

Table~\ref{table:sim} reports the 25\textsuperscript{th}, 50\textsuperscript{th}, and 75\textsuperscript{th} percentile F-scores for \name\ (1st row), specular only (2nd row) and edge only (3rd row). \name's simulator achieves a median F-score of 94\%, while the specular-only or edge-only simulators achieve median F-scores of 90\% and 71\%, respectively. 
An even larger improvement can be seen in the 25\textsuperscript{th} percentile (85\% vs 68\% and 49\%).
This shows the value of \name's techniques for combining multiple simulation types. An analysis of how simulator accuracy varies with $\tau_F$ is in the supplementary.

\vspace{-0.2pt}
\section{Non-Line-of-Sight Applications} 
\label{sec:tasks}
\vspace{-0.1pt}

In this section, we evaluate the ability of 
our dataset and simulator
to enable and establish baselines for NLOS perception for diverse, everyday objects. We describe our experimental setup and analyze the evaluation metrics for NLOS segmentation and classification. Such tasks would be useful in a variety of applications, such as confirming customer orders are properly packaged in e-commerce warehouses, detecting broken objects during shipping, and efficiently finding objects for robotic search.

\vspace{-0.1pt}
\subsection{Object Segmentation}\label{sec:seg}\label{sec:segmentation}
\vspace{-0.1pt}

Here, we show how \name\ can build a mmWave object segmentation tool that works in LOS and NLOS.  
\textbf{Setup.} We build on the state-of-the-art segment-anything model (SAM) \cite{sam}, which takes as input an image to segment and a prompt, either in the form of points in the image or a bounding box. The model returns one or multiple masks, and a score associated with the predicted mask quality. 

Directly using an off-the-shelf object segmentation \hbox{network} for mmWave images faces two challenges. First, SAM is trained on 2D RGB images, but mmWave images are single-channel, complex-valued, 
and 3D. Second, the SAM model requires a prompt before returning segmentation masks. We note that while there are fully automatic methods using SAM (e.g., iterating through a grid of points as a prompt)\cite{sam}, they aim to provide all masks within an image, and would not be applicable in our case since they do not identify which mask contains the object of interest.

\begin{figure*}
\begin{minipage}[t]{0.3\linewidth}
\centering
    \includegraphics[width=0.9\linewidth]{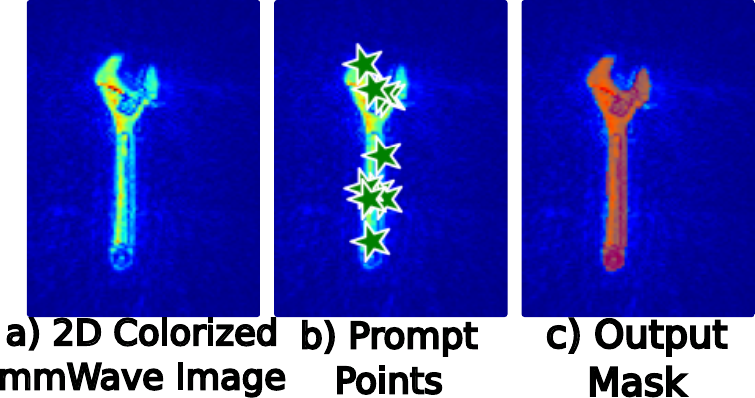}
    \vspace{-0.1in}
    \caption{\footnotesize{\textbf{Segmentation Pipeline.}} \textnormal{a) Colorize 2D image, b) Select prompt points, c) Segment with SAM.}}
    \label{fig:segmentation}
    \vspace{-0.2in}
    % \vspace{-0.3in}
\end{minipage}
\hfill
\begin{minipage}[t]{0.69\linewidth}
\centering
\vspace{-0.8in}
\footnotesize
\label{table:seg}
\setlength{\tabcolsep}{2pt}
\begin{tabular}[t]{ |c V{3} c|c|c|c|c V{3} c|c|c|c|c| }
 \hline  &\multicolumn{5}{c V{3}}{\textbf{LOS}} & \multicolumn{5}{c|}{\textbf{NLOS}}      \\   \cline{2-11}
  \textbf{Input} &\textbf{Pre.} &\textbf{Recall} &\textbf{F1} & \textbf{IoU} & \textbf{Acc.} &\textbf{Pre.}&\textbf{Recall} &\textbf{F1}  & \textbf{IoU} & \textbf{Acc.}    \\  
    \Xhline{3.5\arrayrulewidth} %\hline \hline
Camera   & \cellcolor{Green!50}\textbf{99\%}  & \cellcolor{Green!50}\textbf{99\%} & \cellcolor{Green!50}\textbf{99\%} & \cellcolor{Green!50}\textbf{98\%} & \cellcolor{Green!50}\textbf{99\%} & 
                           \cellcolor{Red!50}3\% & \cellcolor{Green!50}100\%* & \cellcolor{Red!50}7\%   & \cellcolor{Red!50}3\%  & \cellcolor{Red!50}14\%   \\  \hline %hline \hline
24~GHz    & \cellcolor{BurntOrange!25}57\% & \cellcolor{BurntOrange!25}50\% &\cellcolor{BurntOrange!25}37\%  &\cellcolor{BurntOrange!25}22\% &\cellcolor{Green!50}93\% & 
            \cellcolor{BurntOrange!25}53\% & \cellcolor{BurntOrange!25}38\% & \cellcolor{BurntOrange!25}29\% &\cellcolor{BurntOrange!25}17\% &\cellcolor{Green!50}92\%\\ \hline
77~GHz    & \cellcolor{Green!50}98\% & \cellcolor{BurntOrange!25}56\%  &\cellcolor{Green!25}71\%  & \cellcolor{BurntOrange!25}55\% & \cellcolor{Green!50}97\% & 
            \cellcolor{Green!50}\textbf{95\%} & \cellcolor{BurntOrange!25}57\% & \cellcolor{Green!25} \textbf{70\%} & \cellcolor{BurntOrange!25} \textbf{54\%} & \cellcolor{Green!50}\textbf{97\%} \\\hline
% Multi     & \cellcolor{Green!50}     & \cellcolor{Green!25}     & \cellcolor{Green!25}  & \cellcolor{BurntOrange!25} & \cellcolor{Green!50}    
            % & \cellcolor{Green!50}              & \cellcolor{Green!25}& \cellcolor{Green!25} & \cellcolor{BurntOrange!25} & \cellcolor{Green!50} \\ 
Multi Spectral  & \cellcolor{Green!50}97\% & \cellcolor{Green!25}62\% & \cellcolor{Green!25}74\%& \cellcolor{BurntOrange!25}59\% & \cellcolor{Green!50}98\% 
            & \cellcolor{Green!50}\textbf{95\%} & \cellcolor{Green!25}\textbf{63\%} & \cellcolor{Green!25}\textbf{70\%} & \cellcolor{BurntOrange!25}\textbf{54\%} & \cellcolor{Green!50}\textbf{97\%} \\\hline
\end{tabular}
    \vspace{-0.1in}
\captionof{table}{\footnotesize{\textbf{Segmentation Accuracy.} \textnormal{Precision (Pre), Recall, F1 Score, IoU, and Accuracy (Acc) when segmenting camera, 24~GHz, 77~GHz, and multi-spectral images, in both LOS and NLOS scenarios.}} }
    \vspace{-0.2in}
\label{table:seg}
\end{minipage}
% \vspace{-0.15in}
\end{figure*}

To convert our mmWave images to 2D RGB images, we start by averaging the magnitudes along the depth of the 3D image, and use a rainbow colormap for colorization, as shown in Fig.~\ref{fig:segmentation}a. Second, we randomly select a subset of the highest power reflection points as prompts, as shown in Fig.~\ref{fig:segmentation}b. By feeding the resulting RGB image and prompts to the SAM network, we can obtain mmWave segmentations (Fig.~\ref{fig:segmentation}c). We apply the same approach for 24~GHz and 77~GHz images. Additional details can be found in the supplementary.

\noindent \textbf{Multi-Spectral Images.} Beyond segmenting 77~GHz and 24~GHz images separately, we also propose a method for fusing the two images to improve overall segmentation accuracy. To do so, we combine the images during prompt point selection: instead of using a power threshold based only on a single image, the points must exceed thresholds in both 24~GHz and 77~GHz images. Then, we use these points and the 77~GHz colorized image as input to the SAM network. This allows the 24~GHz image to act as a filter, removing the impact of noise (e.g., from occluders) on the point selection and improving segmentation quality.

\subsubsection{Experiments}
\textbf{Baseline.} We compare our segmentation performance to an RGB-D camera baseline. We use the camera image from the experiment directly as input to the SAM model. To select prompt points, we use the depth image to automatically filter out pixels greater than a certain depth (i.e., the background), and randomly select prompt points from the remaining pixels. In the LOS scenarios, this corresponds to points on the object. In NLOS scenarios, this corresponds to points on the occluding object (i.e., the cardboard).

\noindent \textbf{Quantitative Results.}
Tab.~\ref{table:seg} shows the median Precision, Recall, F-score, Intersection-over-Union (IoU), and Accuracy when using the camera, 24~GHz, 77~GHz, and multi-spectral (24~GHz \& 77~GHz) images for both LOS and NLOS scenarios. In LOS, segmenting camera images achieves a median precision, recall and F-score of 99.7\%, 99.5\%, and 99.4\%, respectively. However, in NLOS, the camera only achieves 3.6\% precision and 7.0\% F-score, which is expected since cameras (and other visible-light-based sensors) are unable to operate through occlusions. We note that the camera still achieves 100\% recall since SAM segments the cardboard occluder, covering the full object.

In contrast, the various mmWave image inputs achieve better performance NLOS scenarios, demonstrating how \name\ enables NLOS object segmentation. In particular, the F-score of NLOS for 24~GHz, 77~GHz, and multi-spectral exceeds that of the camera. Notably, multi-spectral images achieve the highest performance. When comparing LOS and NLOS, multi-spectral images only experience a 2.0\%, 0.5\%, and 3.6\% change in precision, recall, and F-score, respectively. This demonstrates that \name\ is able to accurately image and perceive fully occluded objects with a similar performance as in LOS.

Interestingly, the recall when segmenting mmWave images is lower (63.1\%) than the precision (95.1\%). This is primarily because mmWaves are specular, as described earlier, preventing certain portions of the object from appearing in the image. We provide a more detailed discussion on this phenomenon in the supplementary and Sec. \ref{sec:discussion} highlight how future work can improve recall via shape completion.

The IoU and accuracy show similar trends. Notably, 24~GHz, 77~GHz, and multi-spectral all outperform the camera in NLOS. We note that the accuracy is high because it is dominated by the background.

\noindent \textbf{Qualitative Results.} In some cases, the depth camera is unable to measure the depth to every point on the object. This is typically the case for very reflective objects (e.g., mirror-like metallic objects) or clear objects (e.g., glass or plastic). In these cases, \name\ is able to reconstruct more of the object than the depth camera, as shown in Fig.~\ref{fig:res_shiny}. In Fig.~\ref{fig:res_shiny}b, the image from the depth camera has holes due to the object's material reflectivity, whereas in Fig.~\ref{fig:res_shiny}c, \name's mmWave image captures the whole object. Fig.~\ref{fig:res_shiny}d shows our 77~GHz mmWave segmentation masks the entire object.

\vspace{-0.075in}
\subsection{Sim2Real Shape Classification}
\vspace{-0.075in}

Next, we demonstrate how \name\ can be used to classify objects in NLOS based on their geometry. 

% \vspace{-0.1in}
\noindent \textbf{Setup.} We design a classifier which takes 77~GHz mmWave \hbox{images} (after applying the segmentation mask from Sec.~\ref{sec:segmentation}) as input, and outputs the estimated class label. The classifier is a neural network consisting of 5 convolutional layers, followed by 3 fully connected layers, and a softmax activation. We \textit{train the classifier entirely on synthetic data}, and \textit{evaluate on real-world} images in both LOS and NLOS. To generate synthetic images for training, we leverage object meshes provided by YCB dataset.

One challenge in training from synthetic images is to determine which types of reflections dominate for a given object. Recall from Sec.~\ref{sec:simulation} that our simulator models two different reflection types. However, it is difficult to accurately measure the material/texture properties of all objects to identify how to correctly combine these simulation images. Instead, we introduce a new data augmentation, where we choose random weights for each simulation type and sum the weighted images. This allows the network to train on different material properties for each object. We also apply standard image augmentations (rotation, translation, masking, blur). More training details are in supplementary.

\begin{figure*}
\begin{minipage}[t]{0.6\linewidth}
    \centering
    \includegraphics[width=\linewidth]{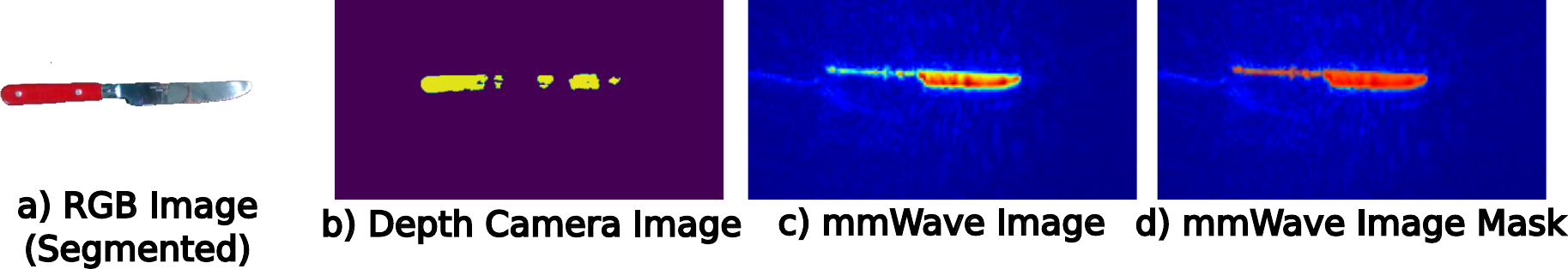}
\vspace{-0.25in}
    \caption{\footnotesize{\textbf{Segmentation Comparison.}} \textnormal{A visual example where the depth image is unable to capture the entire object, but the mmWave image does.} }
\vspace{-0.2in}
    \label{fig:res_shiny}
    \end{minipage}
    \hfill
\begin{minipage}[t]{0.37\linewidth}
\footnotesize
\centering
\setlength{\tabcolsep}{4pt}
\begin{tabular}[b]{ |c|c||c|c|c|}
\hline 
 \multicolumn{2}{|c||}{Training Data} & \multicolumn{3}{c|}{Accuracy}\\
   \Xhline{5\arrayrulewidth}  
  \textbf{Edge}  & \textbf{Spec}   & \textbf{LOS}  & \textbf{NLOS}  & \textbf{Overall}     \\  
 \Xhline{5\arrayrulewidth}  
  X & &  61.1\% & 75.0\% &   67.6\% \\ 
    \hline 
  & X & 77.7\% &  75.0\% &   76.4\% \\  
   \hline 
 X & X &  \textbf{88.8\%} &  \textbf{81.2\%} &   \textbf{85.2\%} \\ 
   \hline 
\end{tabular}
\vspace{-0.1in}
\captionof{table}{\footnotesize{\textbf{Classifier Accuracy} \textnormal{for LOS, NLOS, and overall using specular (Spec), edge, and combined reflection simulations for training. }}}
\label{table:classifier}
    \vspace{-0.2in}
\end{minipage}
\end{figure*}

% \vspace{-0.1pt}
\noindent \textbf{Experiments. } We report the shape classification results on real-world data in LOS and NLOS in Tab.~\ref{table:classifier}. Our classifier performs with an overall accuracy of 85.2\%. Specifically, it achieves an accuracy of 88.8\% and 81.2\% for LOS and NLOS scenarios, respectively, showing that \name\ can be used to enable novel capabilities such as NLOS shape classification.

In Tab.~\ref{table:classifier}, we also include results using a classifier trained on either only edge images (i.e., without combining them with the specular images) or only specular images in order to evaluate the impact of our different simulation types on the classifier accuracy. We observe that these models are only able to achieve 67.6\% and 76.4\%, whereas using both simulation types together achieves 85.2\% accuracy. This shows the benefit of our simulator's different outputs, and our data augmentation technique to combine them.

These results shows that \name\ has the potential to be used for novel NLOS applications, such as confirming packaged orders in closed boxes. Furthermore, by training on synthetic images and evaluating on real-world data, we show that our simulator is representative of the real-world.

% \vspace{-0.1in}
\subsection{Additional NLOS Applications \& Extensions}\label{sec:discussion} 
% \vspace{-0.05in}

In addition to the above applications, \name\ also paves the way for future work in multiple directions:
\begin{itemize}
    \item \textit{NLOS Object Completion:} In some cases, mmWave images do not capture the entire object due to specularity (see supplementary for more details).  It would be interesting future work to create an object completion network that learns the missing parts of an object and creates more photo-realistic NLOS images.
    \item \textit{NLOS 6DoF Estimation:} Another interesting problem is 6DoF estimation from mmWave images. This could be used to ensure proper liquid packaging to avoid spills, or enable efficient grasp planning in NLOS (e.g., grasping an item under cloth or within a box with packing peanuts).
    \item \textit{NLOS Robotic Manipulation: }Future work can leverage many of the above NLOS perception tasks to enable new tasks in robotic manipulation. For example, accurate mmWave object segmentation and classification can significantly increase the efficiency of mechanical search and retrieval (i.e., the problem of searching for a specific item in clutter).  Further, mmWave images can be used to analyze pile stability to improve push and grasp planners. NLOS perception can also help autonomous robots determine which packages contain fragile materials to be handled with special care. More generally, NLOS perception can help autonomous robots to operate in more realistic, cluttered environments.
\end{itemize}

\vspace{-0.05in}
\section{Conclusion}
\vspace{-0.05in}

We present \name, the first millimeter-wave (mmWave) dataset of diverse, everyday objects in both line-of-sight and fully-occluded settings. We collect over 24 million frames of mmWave images, and used them to synthesize 550 high-resolution real world 3D (SAR) mmWave images for 76 objects. We also develop an open-source simulator that can be used to generate synthetic images for any 3D mesh. We demonstrate the utility of our contributions in two applications -- NLOS object segmentation and NLOS shape classification -- establishing baselines for future research and paving the way for important applications in robotic perception, logistics, scene understanding.

As the research evolves, it would be interesting to expand the dataset (e.g., incorporating additional items, surfaces, occlusion types, etc.) which would continue improving the accuracy of the demonstrated NLOS tasks and unlock additional ones. More generally, we hope this work serves as a foundation and motivation for future research in mmWave-based NLOS perception.

% \bibliographystyle{ieeenat_fullname}
% \bibliography{ourbib}

% \bibliographystyle{ieeenat_fullname}
\bibliographystyle{plainnat}
\bibliography{ourbib}

%%%%%%%%%%%%%%%%%%%%%%%%%%%%%%%%%%%%%%%%%%%%%%%%%%%%%%%%%%%%

\appendix

\section{Code and Dataset}
The code and dataset are released here: \url{https://github.com/signalkinetics/MITO_Codebase}

% \section{Video Demonstration}

% % A video demonstrating \name\ is available in the GitHub repo linked above. 

% An anonymized video demonstrating MITO can be found here: 

% \textcolor{blue}{\href{https://www.youtube.com/watch?v=atAqefGl3Sc}{https://www.youtube.com/watch?v=atAqefGl3Sc}}

% \begin{equation}
%     S(r, \theta_A, \theta_D) = \sum_{u=0}^{N} \left(\sum_{v=0}^M \left(\sum_{t=0}^{T} s(i,k,t) e^{j 2 \pi k r t / c} \right)e^{j 2 \pi d v \sin(\theta_A)} \right) e^{j 2 \pi d u \sin(\theta_D)} \notag 
% \end{equation}

\section{Object Diversity}
Since our dataset represents the vast majority of the YCB dataset, our objects are diverse across shape, size, material, and type. In terms of shape, our dataset includes spherical, flat, curved, and complex shapes. In terms of size, it includes objects whose dimensions range from 1.5cm to 47cm. Our objects include various materials such as wood, plastic (e.g., HDPE, PETE, Polypropelene, Nylon, etc), metal, cardboard, rubber, as well as objects with multiple materials. Additionally, our objects represent a wide range of categories, including household tools (wrench, screwdriver, hammer, etc.), groceries (sugar box, soup can, cracker box, etc.), kitchen supplies (spatula, plate, skillet, etc.), toys (marbles, rubik's cube, baseball, etc.), and office supplies (scissors, dry erase markers, etc.).

\section {Why other NLOS imaging techniques are not applicable?}
Alternative NLOS sensing technologies face significant limitations for our scenario. Ultrasound cannot image through closed boxes due to the high acoustic impedance mismatch between air and common occlusion materials (e.g., cardboard). In medical applications, gels are used on Ultrasound transducers to reduce this mismatch with human tissue, which is not practical for object imaging. X-ray signals, while capable of penetrating occlusions, involve ionizing radiation and are therefore unsafe for extended human exposure—unlike mmWave signals, which are non-ionizing and comparable to Wi-Fi. Finally, thermal imaging detects heat sources in NLOS conditions, but this capability does not apply here since the objects of interest do not emit significant heat.

\section{Sample Dataset Images}%\textred{Visualizing \name}}

Here, we include sample mmWave images for 50 out of the 76 objects in \name.  Note that the majority of our objects contain multiple real-world images from different angles or measurments apertures. 

Fig.~\ref{fig:all_images} shows the segmented RGB images (1st column), 77GHz line-of-sight (2nd column), 77GHz non-line-of-sight (3rd column), 24GHz line-of-sight (4th column), and 24GHz non-line-of-sight (5th column) mmWave images from real-world experiments.  The objects are sorted by their unique identifier (defined by the YCB dataset). This shows the diversity of objects within \name, as well as the performance of the mmWave imaging for different objects. 

\section{Impact of Occlusion types}

\begin{figure}
\centering
    \includegraphics[width=0.99\linewidth]{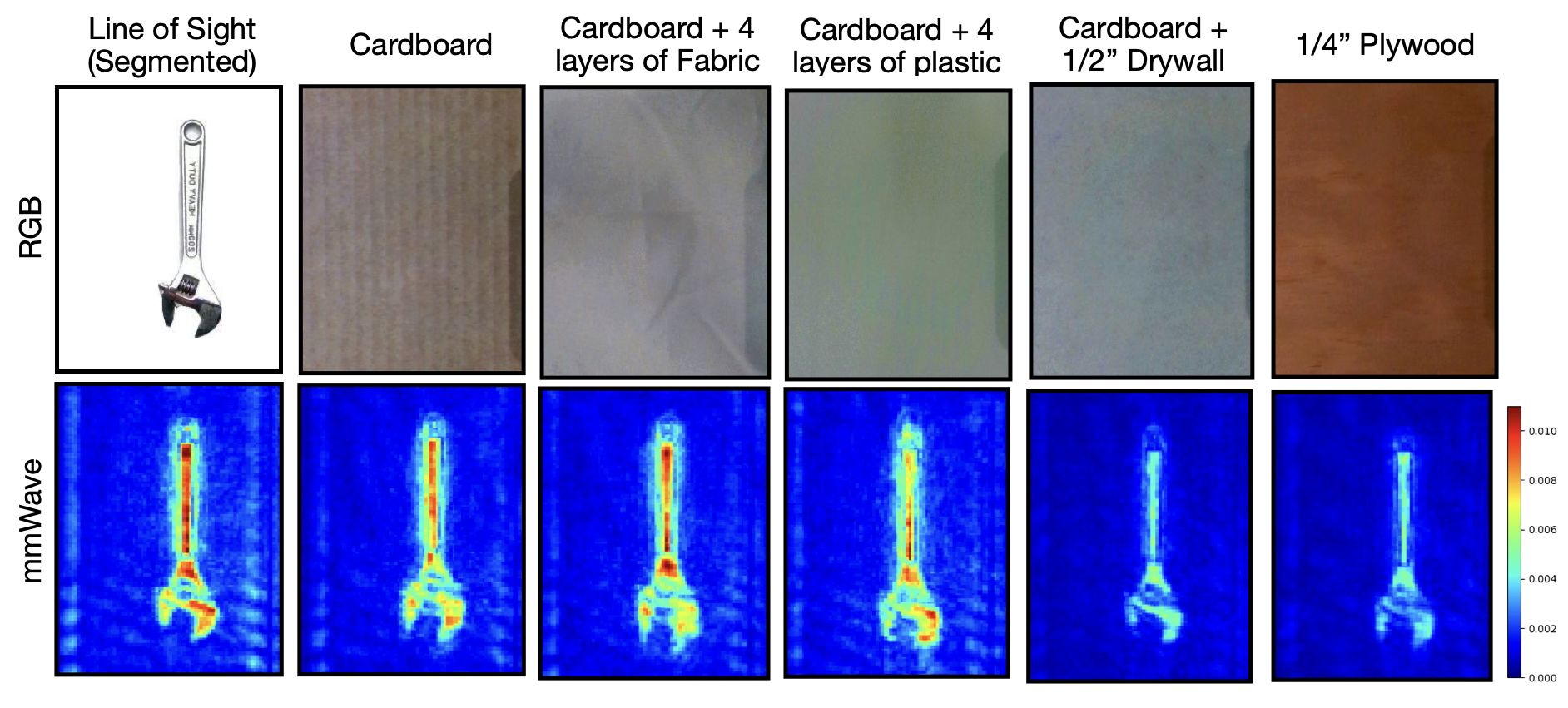}
    \vspace{-0.1in}
    \caption{\footnotesize{Occlusions Materials impact on mmWave}} 
    \label{fig:occlusions}
    \vspace{-0.15in}
\end{figure}

Different occlusion materials can influence the mmWave signals propagation and thus the quality of mmWave images. Prior wireless research has studied how different materials attenuate RF signals~\cite{60ghz_study}, often modeling this effect via changes in signal-to-noise ratio. Thus, we expect different occlusions to primarily change the brightness or noisiness of our mmWave images.\footnote{Metal obstacles may block all wireless signals and prevent imaging.} We verify this through a qualitative analysis in Fig.~\ref{fig:occlusions}, which shows the RGB and mmWave images of a metallic wrench. It compares LOS to multiple occlusion types (cardboard, fabric, plastic, drywall, \& plywood), plotting all heatmaps with the same color scale.\footnote{Our occlusions are 1 layer of cardboard, 1 layer of cardboard plus 4 layers of a fabric sheet, 1 layer of cardboard plus four layers of a plastic shower curtain, 1 layer of cardboard plus 1/2’’ drywall, and 1 sheet of 1/4’’ plywood.} We note that:

\begin{itemize}
    \item The wrench is clearly visible in all images, showing that mmWave imaging works through various occlusions.
    \item The intensity of the wrench is lower for the drywall and wood (note that the images do not contain any red regions), which is expected since these materials introduce greater attenuation.
    \item Some occlusions show increased noise around the wrench, especially for the deformable plastic and fabric occlusions. These occlusions do not lie completely flat, and thus can induce small distortions in the image. This highlights the need for models and algorithms which can account for mmWave-specific behavior to enable NLOS perception tasks. Such occlusions would be valuable additions for future extensions of MITO.
\end{itemize}

\section{Aligning RGB-D and mmWave Data}

Many perception tasks require ground-truth knowledge of an object's shape and location for training and evaluation. For example, our segmentation task in Sec. 6.1 of the main paper requires ground-truth 2D binary masks of the object for evaluation. Other tasks, such as surface reconstruction, 6D pose estimation, and 3D segmentation, would require a ground-truth 3D point cloud of the object's surface.

In this section, we describe how we obtain ground-truth object point clouds and segmentation masks from RGB-D data, and align these with the 3D mmWave image. We leverage RGB-D images captured from a wrist mounted camera at the beginning of each experiment (when the object is still in line-of-sight). First, to isolate the object within the RGB-D image, we manually segment the object within the RGB image (see Sec. 4.3 in the main paper for more details). Then, we apply the following four steps to align the segmented RGB-D data with the mmWave image and produce both ground-truth 2D segmentation masks and 3D point clouds:

\noindent
\textit{1. Transforming mmWave data}: First, we use the radar's poses during the mmWave image acquisition to transform the 3D mmWave images to the world frame:

{\small
\vspace{-0.12in}
\begin{equation} \label{eq:mm_world}
~^W\!P^U= ~^W\!R^M \times ~^M\!P^U +   ~^W\!P^M  
\vspace{-0.035in}
\end{equation}}

\noindent
where {\small$~^W\!P^U$} and {\small$~^M\!P^U$} are the position (x,y,z) of a voxel {\small$U$} in the mmWave image in the world frame {\small$W$} and mmWave radar frame {\small$M$}, respectively; {\small$~^W\!P^M$} and {\small$~^W\!R^M$} are the position and rotation of the radar frame {\small$M$} in the world frame. 
% \note{Note to Tara: I (Laura) am a little confused. What is {\small$~^H\!P^M$}? }

\noindent \textit{2. Transforming Visual Data: } Next, we use the RGB-D data to produce a 3D point cloud of the object in the world frame. To do so, we apply the manually segmented RGB mask to the camera's depth image.\footnote{The realsense depth camera provides aligned RGB and depth images.} Then, we convert each point within the segmented depth image into a 3D point  using the camera's intrinsic properties. Finally, we use the camera's pose to convert this point cloud from the camera's frame to the world frame:

{\small
\vspace{-0.12in}
\begin{equation} \label{eq:rgb_world}
~^W\!P^V= ~^W\!R^C \times ~^C\!P^V +   ~^W\!P^C 
\vspace{-0.03in}
\end{equation}}

\noindent
where {\small$~^W\!P^V$} and {\small$~^C\!P^V$} are a point (x,y,z) from vision-based point cloud in the world frame $W$ and camera frame $C$, respectively; {\small$~^W\!P^C$} and {\small$~^W\!R^C$} are the position and rotation of the camera frame in the world frame, respectively.  

This point cloud is now in the same frame as the mmWave image in the world frame ($~^W\!P^U$ in Eq.~\ref{eq:mm_world}) and can serve as 3D ground truth for different perception tasks.

% \textblue{This point cloud is now in the same frame as the 3D mmWave image ($~^W\!P^U$ in Eq.~\ref{eq:mm_world}), and can serve as 3D ground truth for different perception tasks such as pose estimation.}

\begin{figure}
\centering
    \includegraphics[width=0.99\linewidth]{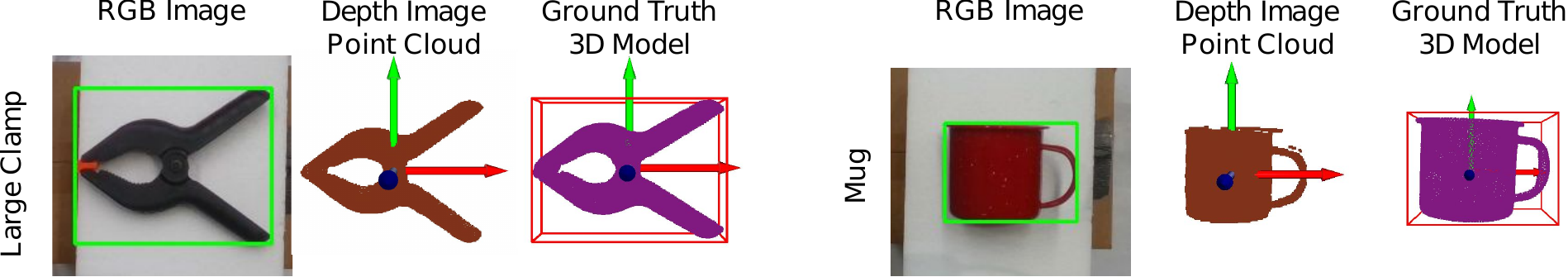}
    \vspace{-0.1in}
    \caption{\footnotesize{{Example dataset annotations for two objects (plastic clamp and mug). Each object shows different modalities and annotation types: RGB image with 2D bounding box, depth image point cloud, and ground truth 3D mesh with 6-DoF pose and bounding box.}}} 
    \label{fig:dof}
    \vspace{-0.15in}
\end{figure}

\noindent
\textit{3. Constructing 2D Binary Masks:} Next, we produce ground-truth binary masks. To do so, we voxelize the vision-based point cloud ($~^W\!P^V$ in Eq.~\ref{eq:rgb_world}). Here, each voxel's center and size is the same as that of the mmWave image's voxels. Then, we project both the 3D vision-based image and 3D mmWave image into two dimensions. Now, the projected 2D vision-based image is a binary mask which acts as a ground-truth segmentation for the 2D mmWave image. {It can also be used to generate a 2D bounding box of the object as shown in the RGB images in Fig.~\ref{fig:dof}.}

\noindent \textit{4. Creating 3D Ground-Truth Models.} We also provide ground-truth 3D models and 3D bounding boxes. To do so, we leverage meshes from the YCB dataset for each object, and convert them into point clouds by uniformly sampling points along the surface. Then, we align these point clouds to the vision-based point clouds in the world frame ($~^W\!P^V$). We use the iterative closest point (ICP) algorithm to find the transformation and rotation which best aligns the full 3D point cloud to the measured depth information. This provides a ground-truth 3D representation of the object in the world frame, which can also be used to derive a 3D bounding box~\footnote{We find this bounding box using Open3D's \texttt{get\_axis\_aligned\_bounding\_box} function.} We show examples of this annotation in Fig.~\ref{fig:dof}.

\section{Radar \& SAR Details:}
Our radars leverage frequency-modulated continuous wave (FMCW) signals, which span a large bandwidth. The bandwidths are 3~GHz for the 77~GHz radar and 200~MHz for the 24~GHz radars, respectively.

 Given the bandwidth and aperture, we can compute \name's resulting resolution. \name's resolution in the X,Y, and Z dimensions (computed using Eqs. 2-3 in the main paper) are 3.1~mm, 1.5~mm, and 4.9~cm for the 77 GHz radar and 1~cm, 0.5~cm, and 0.75~m for the 24 GHz radar, respectively. 
 % \textcolor{red}{add sentence on why xyz?}

% \vspace{-0.05in}
\section{Task 1: Object Segmentation}
% \vspace{-0.03in}
Here, we provide additional details on the object segmentation from Sec 6.1 in the main paper.

\subsection{Additional Design Details}
% Here, we provide additional details about our solution to leverage SAM for mmWave object segmentation. We first need to
We describe how we leverage SAM for mmWave segmentation in more detail. To do so, we first need to
convert our 3D, complex-valued, single-channel mmWave images to 2D RGB images that can be input to the SAM model. We start by projecting our 3D image into a real, 2D image by averaging the magnitudes along the depth of the image. Then, as illustrated in the example in Fig.~7a in the main paper, we colorize the single-channel image using a rainbow colormap, such that high values appear red and low values appear blue. The resulting three-channel image can then be input into SAM.

Next, we need to choose one or more points to prompt the network. One may think that we can simply choose the highest value pixel as the prompt. However, this typically results in an insufficient mask that does cover the entire object. Instead, we choose a number of points that are dispersed throughout the object such that the network is encouraged to segment the entire object. To do so, we first find the set of all points within a fixed percentage of the maximum power, $S_p$:

\vspace{-0.1in}
{\small
\begin{equation}
    S_p = \{ p\ |\  |I(p)| > \tau * \max(|I|) \}
\end{equation}
}

\noindent where $|I(p)|$ is the magnitude of the mmWave image at point p, $\tau$ is a threshold percentage. 

Since this will be a large number of points, we then randomly choose a subset:
%of these points to use for the final prompt.

\vspace{-0.1in}
{\small
\begin{equation}
    P_p = \{ p_i\ |\ p_i \in_R S_{p} \cap i \in \{1,..,K\} \}
\end{equation}
}

\noindent where $P_p$ is the final set of points used to prompt the SAM network, K is the number of chosen points, and $\in_R$ denotes a randomly selecting from a set. An example of these points is shown in Fig.~7b. 

Now, the converted RGB image and prompt points are fed to the SAM network to produce mmWave segmentations. An example  output is shown in Fig.~7c by a red mask over the segmented pixels.  %Additional implementation and evaluation details are in the supplementary materials.

\subsection{Implementation Details}

We implement the SAM model using pre-trained weights provided by Meta. Then, SAM can be used to output multiple possible segmentation masks for a given prompt, along with a score for each  mask. To remove the invalid mmWave masks, we remove any masks that take up a large percentage of the image.  From the remaining masks, we select the mask with the highest score.

\subsection{Evaluation Details}

% \subsubsection{Evaluation Metrics} 
\noindent \textit{Evaluation Metrics:} 
We evaluate our segmentation with precision, recall, F-Score, IoU, and Accuracy. These are defined as:
\begin{equation}
    P = \frac{\texttt{TP}}{\texttt{TP}+\texttt{FP}} \ \ \ \ \ \ \ \  \ \ R = \frac{\texttt{TP}}{\texttt{TP}+\texttt{FN}}
\end{equation}
\begin{equation}
    F = \frac{2\ P \ R}{P+R}  
\end{equation}
\begin{equation}
    IoU = \frac{TP}{TP+FP+FN}
\end{equation}
\begin{equation}
    Acc = \frac{TP+TN}{TP+TN+FN+FP}
\end{equation}
\noindent where $P$, $R$, $F$, $IoU$, and $Acc$ are the precision, recall, F-Score, IoU, and Accuracy, respectively. $\texttt{TP}$, $\texttt{TN}$, $\texttt{FP}$, and $\texttt{FN}$ are the number of true positives, true negatives, false positives, and false negatives, respectively.  

% \textcolor{red}{Add other metrics or cut this}

% \subsubsection{Ground Truth}
\noindent \textit{Ground Truth:}
To evaluate the object segmentation, we use the ground truth masks described in Sec. F.3. Note that the depth image captured by the camera sometimes contains NaNs due to its inability to accurately capture some parts of objects. To resolve this problem, we fill in any missing depth value with the average depth value of the mask before transforming it to the radar frame.

\noindent \textit{Evaluation:} Since our point selections relies on a random process, we measure the above metrics 3 times for each object using a different instance of prompt selection and segmentation. We evaluate our segmentation on a subset of our dataset across 236 real-world images (both 24~GHz and 77~GHz).

\subsection{Analyzing Segmentation Performance}
% \begin{figure}

% \centering
%     \includegraphics[width=\linewidth]{app/figures/orange_overlay_new3.pdf}
%     \vspace{-0.1in}
%     \caption{\footnotesize{\textbf{Recall.}} \textnormal{This figure shows two examples where the mmWave image exhibits high and low recall. a) shows the segmented RGB image for a padlock, and b) shows its corresponding mmWave image, where the full object is captured in the image, and results in high recall. c) shows a second example of the segmented RGB image for a spherical orange. d) is the corresponding mmWave image for this object. Due to specular reflection of mmWave from the orange's surface, only a small portion of this object is captured in the mmWave image.} } 
%     \label{fig:res_orange}
%     \vspace{-0.15in}
% \end{figure}

\begin{figure}
\begin{minipage}[t]{0.7\linewidth}
\centering
    \includegraphics[width=\linewidth]{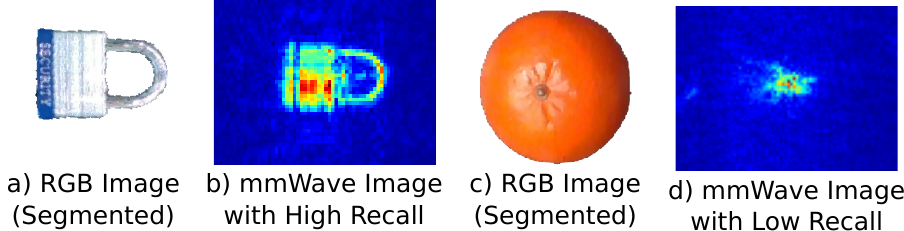}
    \vspace{-0.1in}
    \caption{\footnotesize{\textbf{Recall.}} \textnormal{This figure shows two examples where the mmWave image exhibits high and low recall. a) shows the segmented RGB image for a padlock, and b) shows its corresponding mmWave image, where the full object is captured in the image, and results in high recall. c) shows a second example of the segmented RGB image for a spherical orange. d) is the corresponding mmWave image for this object. Due to specular reflection of mmWave from the orange's surface, only a small portion of this object is captured in the mmWave image.} } 
    \label{fig:res_orange}
    \vspace{-0.15in}
\end{minipage}
\hfill
\begin{minipage}[t]{0.275\linewidth}
\centering
    \includegraphics[width=\linewidth]{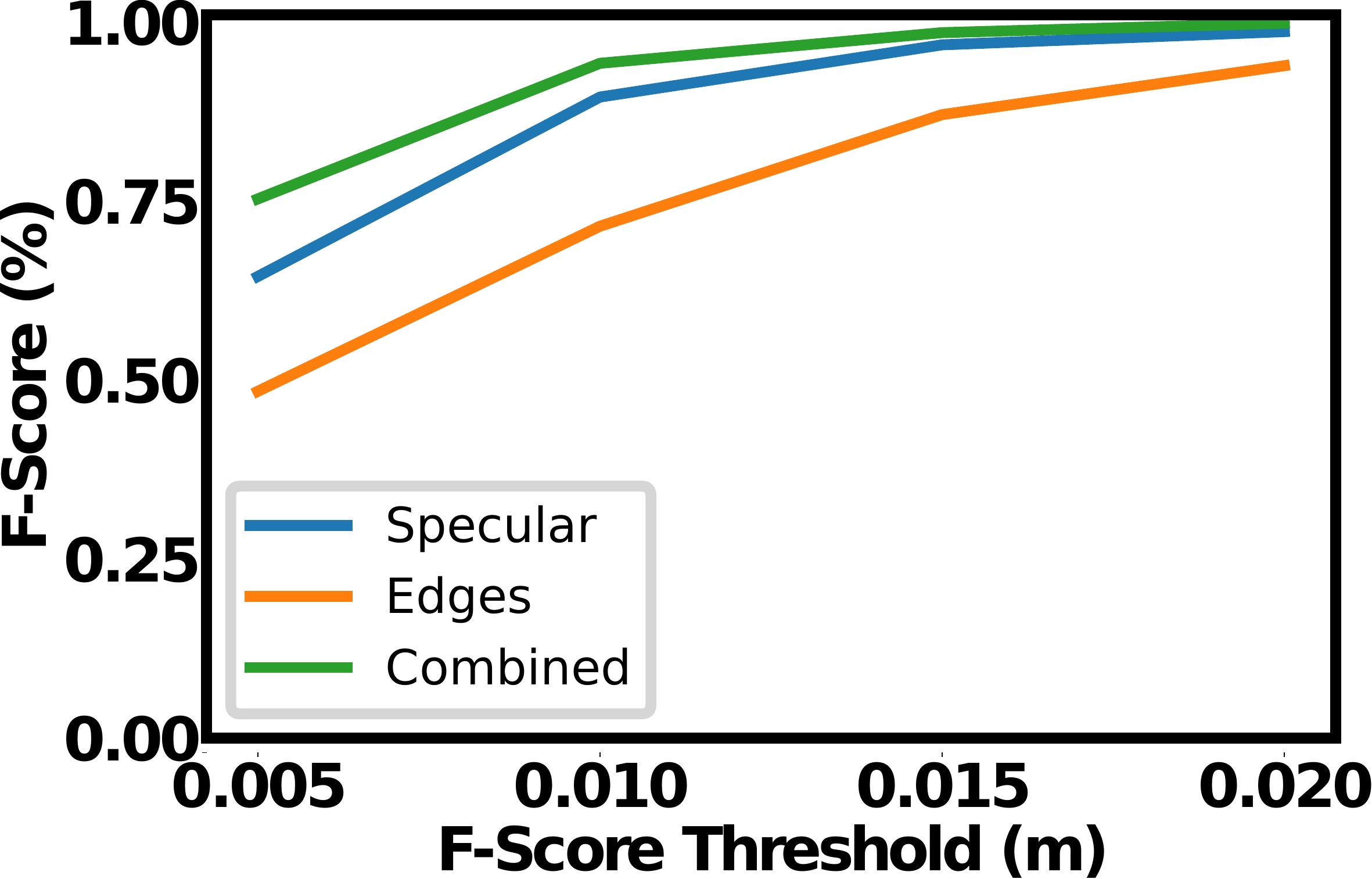}
    \vspace{-0.1in}
    \caption{\footnotesize{Impact of F-Score threshold on simulator accuracy}} 
    \label{fig:fscore_threshold}
    \vspace{-0.15in}
\end{minipage}
\end{figure}

As shown in Sec.~6.1 in the main paper, the median recall of our segmentation {(63\%) is lower than the precision (95\%)}. To understand why, we analyze an example of two different objects in Fig.~\ref{fig:res_orange}. In the first example, Fig.~\ref{fig:res_orange}a shows the RGB image of a padlock. For this object, Fig.~\ref{fig:res_orange}b shows that the mmWave image captures a large portion of the object. This is because the surface is fairly flat. 

In the second example, Fig.~\ref{fig:res_orange}c shows an example of a spherical orange. Fig.~\ref{fig:res_orange}d shows the mmWave image for this object. 
In this case, the object is not fully captured in the mmWave image. This is because, at millimeter-wave frequencies, the object is experiencing specular reflections (see Sec.~3.3 in the main paper). Since the object is spherical, the sides of the object will reflect signals away from the radar, preventing the radar from receiving them and the mmWave imaging algorithm from recovering the full object's shape. This leads to a low segmentation recall performance when compared with the ground-truth segmentation mask produced from the RGB image.

In contrast to mmWave images, visible light  experiences primarily diffuse reflections from the surface of the orange, which results in full recovery of the orange's shape within the camera image. 
This is one example of how mmWaves images are fundamentally different from camera images. Further, it shows the potential benefit for future work on generalizable mmWave object completion.

\section{Task 2: {Sim2Real Shape} Classification}

Here, we provide additional details on the {shape} classification from Sec.~6.2 in the main paper. 
\subsection{Implementation Details}

\noindent \textit{Network Design:} We design a neural network with 5 convolutional layers, followed by 3 fully connected layers, and a softmax activation. The input to the network is a mmWave image (after applying the binary mask generated in Sec. 6.1). The network then outputs the class probabilities. We sort our dataset into 4 classes: {Boxes, Cans, Spheres, and Handle-based tools}\footnote{For the purpose of this classifier, we do not use objects from the dataset that do not fit in these categories when training/testing the classifier.}. 

\noindent \textit{Training: }We train our network on an Ubuntu 22.04 computer with a GTX 1080 GPU. We use an adam optimizer with a learning rate of 1e-3, and we train our network for 500 epochs. Since our classes are unbalanced, we sample our data during training such that each class is equally represented. 

\noindent \textit{Augmentations:} In addition to our mmWave augmentation described in Sec.~6.2, we also implement a number of other randomized augmentations: rotations, translations, scaling, masking, horizontal and vertical flipping. 

\noindent \textit{Data Split: }The training set is comprised entirely of simulation data. Then, the real world data (LOS and NLOS) is randomly distributed into a test and validation set (with equal sizes), such that each class is represented equally in the test and validation set. 

\subsection{Evaluation Details}

\noindent \textit{Evaluation Metrics:}
% \noindent \textbf{Classification Accuracy} 
We evaluate our mmWave classification with the accuracy, defined by the percentage of correctly classified images in the test set.

\noindent \textit{Ground Truth:} For the object classifier, we use the ground truth labels for each experiment and manually sort the objects into the above classes.

\section{Impact of F-Score threshold on simulator accuracy}
In the main paper (Sec~5.2), simulator accuracy is evaluated using the 3D F-score metric with a fixed threshold. This threshold, denoted as $\tau_F$ in Equ. 9 and 10, determines the maximum allowable distance between corresponding points in the real and synthetic point clouds for them to be considered a match. To better understand the sensitivity of simulator performance to this parameter, we analyze how the F-score varies across a range of thresholds from 0.005~m to 0.02~m. Fig.~\ref{fig:fscore_threshold} plots the F-score for all objects in the dataset under three simulation settings: combined simulation (green), specular-only (blue), and edge-only (orange).

We observe that even at a strict threshold of 0.005~m, the combined simulation maintains an F-score above 75\%, indicating strong alignment with real-world data. As the threshold increases, all methods show improved scores, but the combined simulation consistently outperforms the others. We also note that F-Score thresholds $\geq$1~cm are common for mmWave-based systems [18], as they have lower resolutions than their RGB-D counterparts.

% \begin{figure}
% \centering
%     \includegraphics[width=0.8\linewidth]{app/figures/fscore_threshold2.pdf}
%     \vspace{-0.1in}
%     \caption{\footnotesize{Impact of F-Score threshold on simulator accuracy}} 
%     \label{fig:fscore_threshold}
%     \vspace{-0.15in}
% \end{figure}

% \begin{figure*}
% \begin{minipage}[t]{0.47\textwidth}
% \centering
%     \includegraphics[width=1\textwidth]{figures/all_supp.pdf}
%     \vspace{-0.2in}
%     \caption{\footnotesize{\textbf{.}} \textnormal{}}
%     \label{fig:sim}
%     \vspace{-0.2in}
    
% \end{minipage}
% \hfill
% \begin{minipage}[t]{0.47\textwidth}
% \centering
%     \includegraphics[width=1\textwidth]{figures/all_supp.pdf}
%     \vspace{-0.2in}
%     \caption{\footnotesize{\textbf{.}} \textnormal{}}
%     \label{fig:sim}
%     \vspace{-0.2in}
    
% \end{minipage}
% % \centering
% %     \includegraphics[width=0.5\textwidth]{figures/all_supp.pdf}
% %     \vspace{-0.2in}
% %     \caption{\footnotesize{\textbf{.}} \textnormal{}}
% %     \label{fig:sim}
% %     \vspace{-0.2in}
% \end{figure*}

% \bibliographystyle{ieeenat_fullname}
% \bibliography{main}

% \newpage

\begin{figure*}
\centering
    \includegraphics[width=0.9\textwidth]{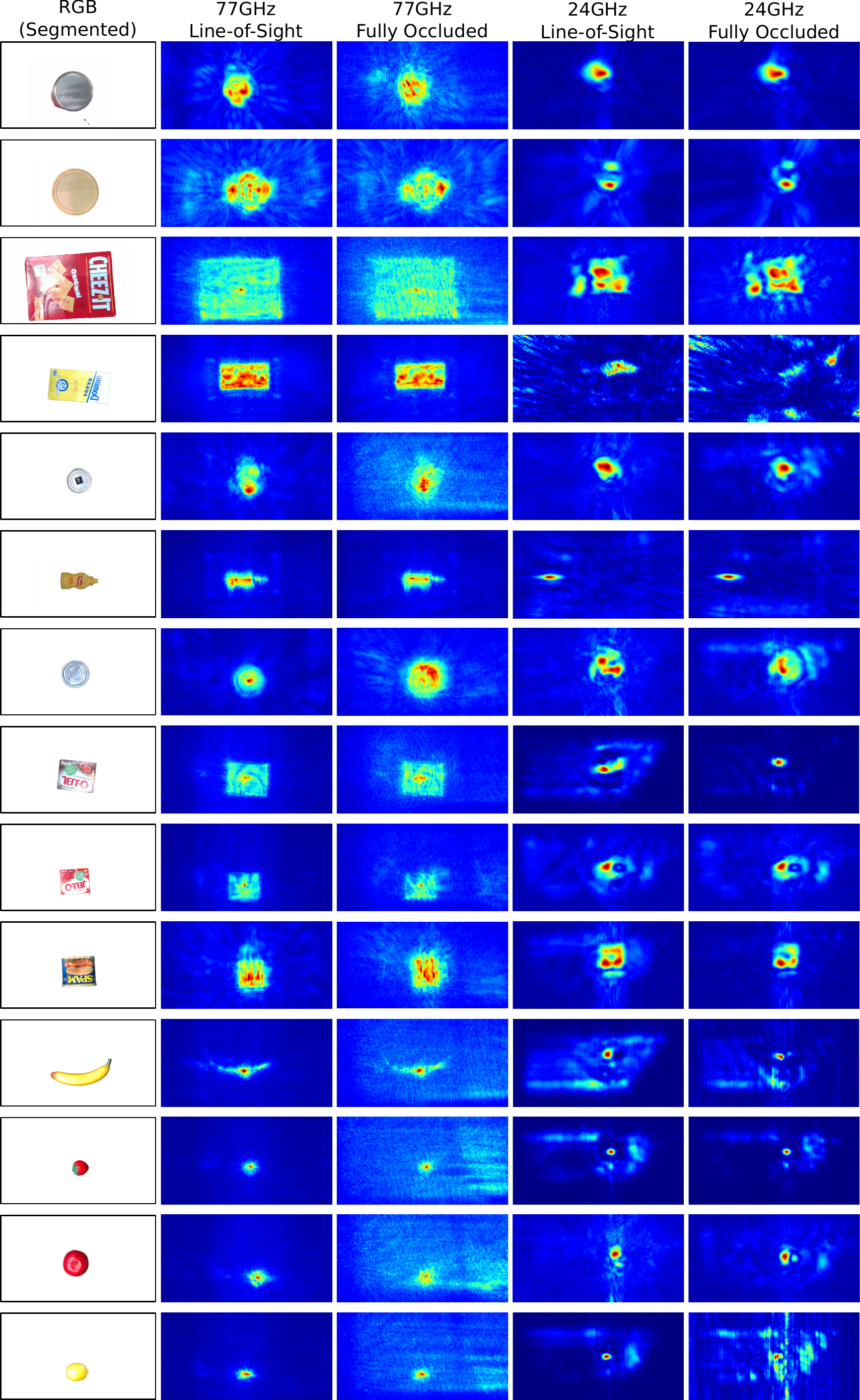}
    \vspace{-0.1in}
    % \caption{\footnotesize{\textbf{Sample Images of \name.}} \textnormal{This figure shows the segmented RGB images (1st column), a sample 77GHz line-of-sight (2nd column) 77GHz non-line-of-sight (3rd column), 24GHz line-of-sight (4th column) and 24GHz non-line-of-sight (5th column) mmWave image from a real-world experiment \textred{for 50 objects within \name}.}}
    % \label{fig:all_images}
    \vspace{-0.2in}
    
\end{figure*}

% \newpage
\begin{figure*}
\centering
    \includegraphics[width=0.9\textwidth]{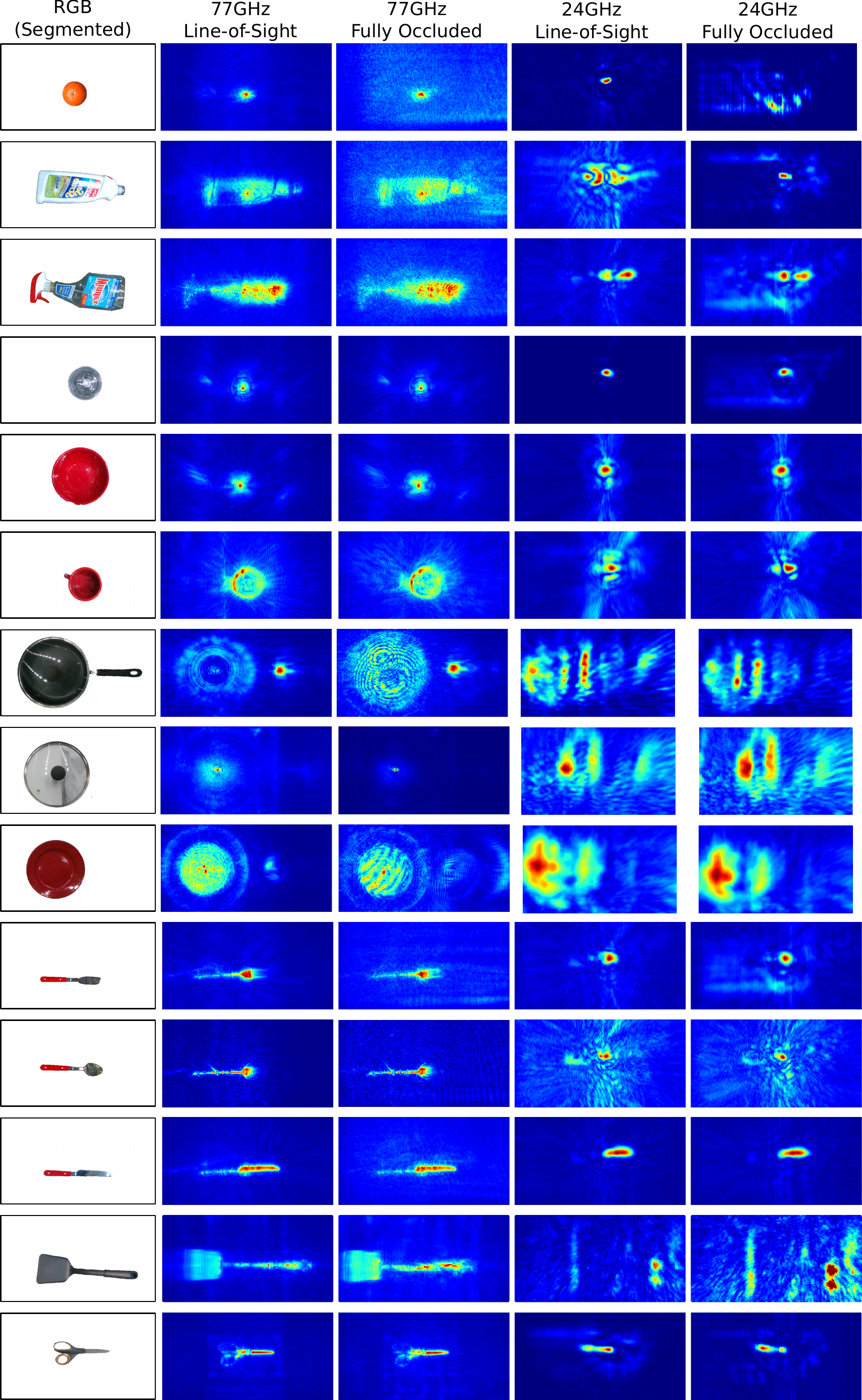}
    \vspace{-0.1in}
    % \caption{\footnotesize{\textbf{Sample Images of \name.}} \textnormal{This figure shows the segmented RGB images (1st column), a sample 77GHz line-of-sight (2nd column) 77GHz non-line-of-sight (3rd column), 24GHz line-of-sight (4th column) and 24GHz non-line-of-sight (5th column) mmWave image from a real-world experiment \textred{for 50 objects within \name}.}}
    % \label{fig:all_images_2}
    % \vspace{-0.2in}
    
\end{figure*}

\begin{figure}
\centering
    \includegraphics[width=0.9\textwidth]{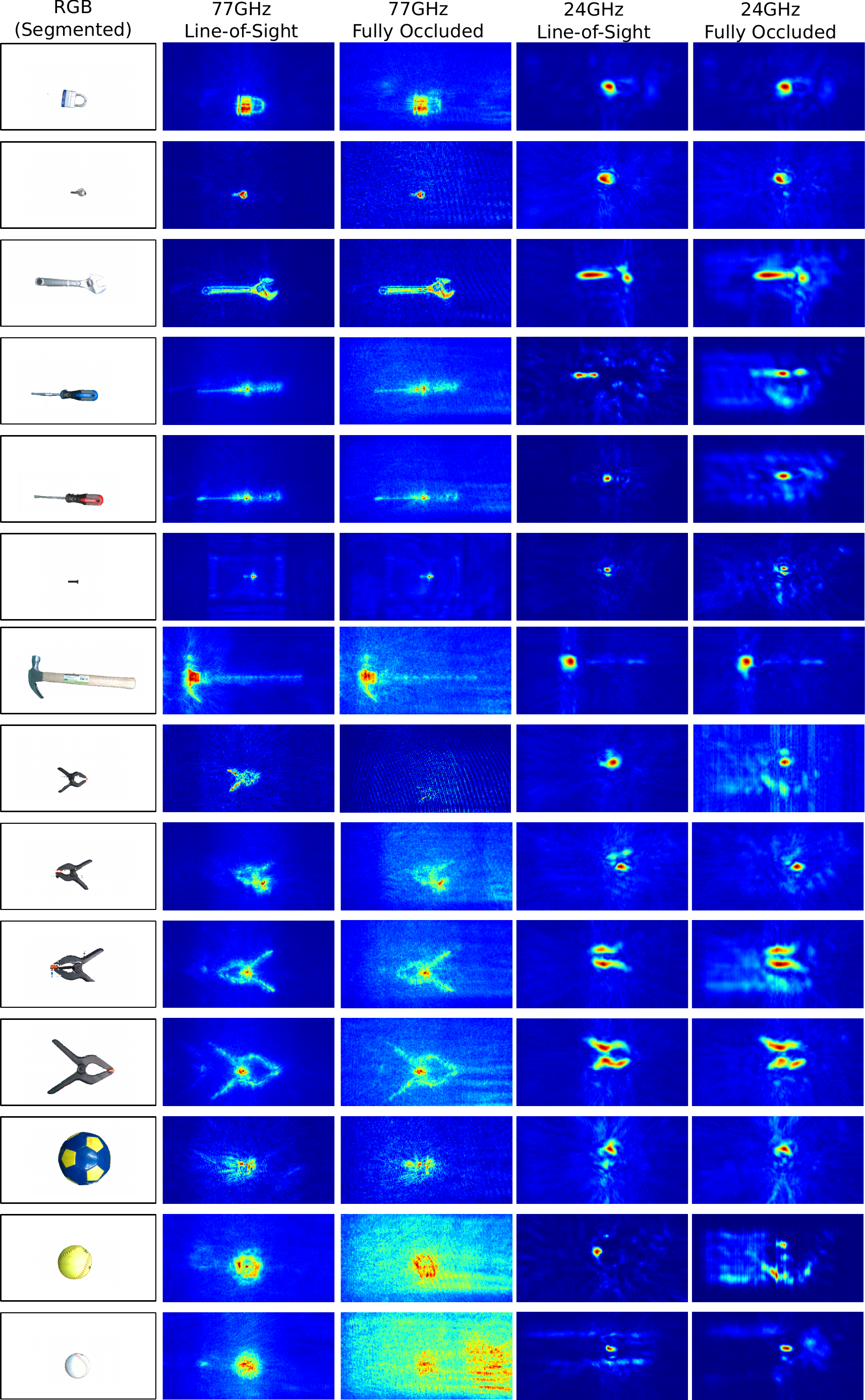}
    % \vspace{-0.1in}
    % \caption{\footnotesize{\textbf{Sample Images of \name.}} \textnormal{This figure shows the segmented RGB images (1st column), a sample 77GHz line-of-sight (2nd column) 77GHz non-line-of-sight (3rd column), 24GHz line-of-sight (4th column) and 24GHz non-line-of-sight (5th column) mmWave image from a real-world experiment \textred{for 50 objects within \name}.}}
    % \label{fig:all_images_3}
    % \vspace{-0.2in}
    
\end{figure}

\begin{figure}[t]
\centering
% \vspace*{-2.5in}
    \includegraphics[width=0.9\textwidth]{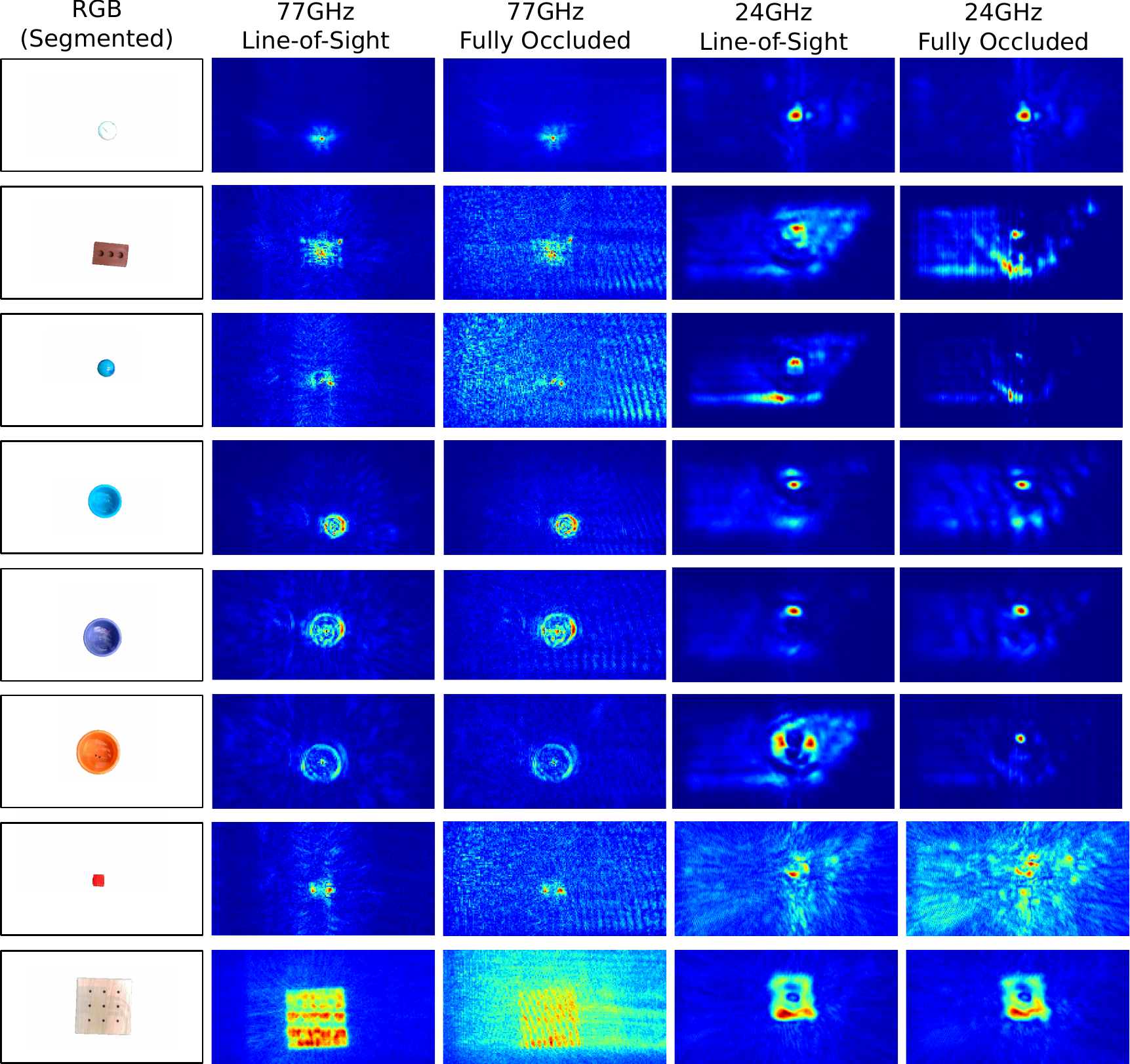}
    \vspace{-0.1in}
    \caption{\footnotesize{\textbf{Sample Images of \name.}} \textnormal{This figure shows the segmented RGB images (1st column), a sample 77GHz line-of-sight (2nd column) 77GHz non-line-of-sight (3rd column), 24GHz line-of-sight (4th column) and 24GHz non-line-of-sight (5th column) mmWave image from a real-world experiment.}}
    \label{fig:all_images}
    % \vspace{-0.2in}
    
\end{figure}

\newpage
~
\newpage

%%%%%%%%%%%%%%%%%%%%%%%%%%%%%%%%%%%%%%%%%%%%%%%%%%%%%%%%%%%%

% \newpage
% \input{checklist.tex}

\end{document}